\documentclass[10pt,twocolumn,letterpaper]{article}

\usepackage[pagenumbers]{wacv}      
\usepackage[accsupp]{axessibility} 
\usepackage{multirow}
\usepackage{algorithm}
\usepackage{algorithmic} 
\usepackage{amsthm}
\usepackage{amssymb}
\usepackage{amsmath}
\usepackage{graphicx}
\usepackage{blindtext}
\usepackage{multicol}
\usepackage{multirow}
\usepackage{caption}
\usepackage{amsthm}

\newtheorem*{remark}{Remark}
\usepackage[table]{xcolor} 
\usepackage{tcolorbox}
\definecolor{softgreen}{HTML}{c8e7ff} 
\usepackage{booktabs}

\newcommand{\tbd}{\textcolor{gray}{--}}
\newcommand{\nax}{\textcolor{gray}{---}}

\usepackage{makecell}

\definecolor{wacvblue}{rgb}{0.21,0.49,0.74}
\usepackage[pagebackref,breaklinks,colorlinks,allcolors=wacvblue]{hyperref}
\def\wacvPaperID{785} 
\def\confName{WACV}
\def\confYear{2027}

\title{Neural Implicit Energy Optimization for LiDAR Mapping}

\author{
Hrishikesh Viswanath$^{1}$,
Md Ashiqur Rahman$^{1}$,
Chi Lin$^{1}$,
Damon Conover$^{2}$,
Aniket Bera$^{1}$\\
$^{1}$Purdue University, USA \quad
$^{2}$DEVCOM, Army Research Laboratory, USA\\
{\tt\small hviswan@purdue.edu}
}

\begin{document}
\twocolumn[{%
\renewcommand\twocolumn[1][]{#1}%
 \maketitle
\begin{center}
    \centering    
\captionsetup{type=figure}
\includegraphics[width=1\linewidth]{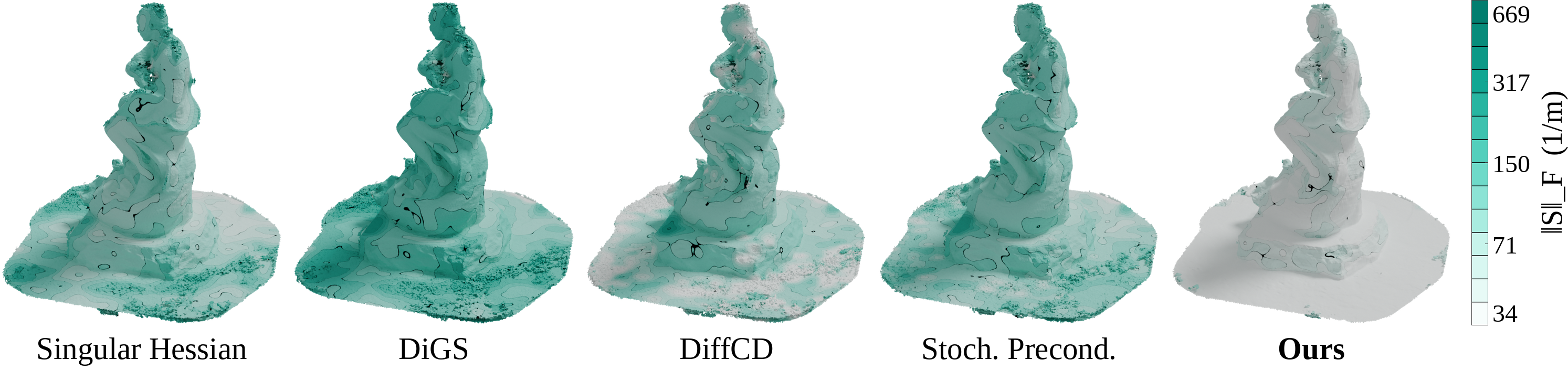}
    \captionof{figure}{\textbf{ Surface Energy Optimized LiDAR maps.} We visualize the spurious features (jitter) of reconstructed LiDAR maps, measured by the shape operator norm $|S|_F$ defined over the surface. Our approach improves the smoothness of the surface while preserving the geometric fidelity of the scene without incurring any additional inference time costs for large scale LiDAR sweeps.}
\label{fig:energy_row}
\end{center}%
}]
\begin{abstract}
Neural implicit representations provide memory-efficient signed distance field (SDF) parameterizations for large-scale LiDAR scenes, but the geometry recovered from sparse and often noisy inputs carries jitter and artifacts. Classical geometric optimization techniques resolve such artifacts with surface energy functionals, whose minimizers correspond to smooth surfaces. Motivated by this principle, we formulate neural LiDAR mapping as the optimization of energy functionals over the parameter space of neural implicit representations. Specifically, we regularize the signed distance field with an energy functional defined on the shape operator of its level set, which represents the curvature of the surface, thus encouraging smoothness along the surface. This adapts classical geometric priors to the scale of million-point LiDAR maps. Experiments on large-scale datasets show reconstructions with improved smoothness at preserved geometric fidelity, and no additional inference-time cost. The code is publicly available at \href{https://github.com/HrishikeshVish/nieo}{https://github.com/HrishikeshVish/nieo}
\end{abstract}
    
\section{Introduction}
\label{sec:intro}

Constructing accurate geometric representations of large-scale environments is a fundamental task in 3D computer vision. These spatial models serve as the foundation for downstream visual reasoning tasks, simulation frameworks, and autonomous systems \citep{aulinas2008slam,macario2022comprehensive,andronie2022remote,lluvia2021active,senthilkumaran2023artemis}. To capture these environments, cameras and LiDAR (Light Detection and Ranging) are the two most common sensors utilized in practice. While cameras capture dense photometric information, LiDAR provides the depth perception essential for spatial modeling and geometric mapping. However,  relying directly on raw LiDAR scans presents challenges. Captured point clouds can be noisy and often sparse, particularly in areas with limited sensor coverage. This sparsity makes generating high-fidelity surfaces complex, and the resulting mapped surfaces often exhibit jitter. Ensuring that the reconstructed surface accurately represents the continuous underlying geometry while preserving localized details remains a challenge in large-scale 3D mapping.

Neural implicit 3D reconstruction frameworks typically map scenes using learnable continuous feature representations \cite{williams2022neural, huang2023neural}. When handling large-scale geometry, these frameworks represent the environment using sparse hierarchical spatial features. This serves as a differentiable latent state, using which an implicit neural representation learns a continuous signed distance field (SDF)~\citep{zhong2023shine, sun20243qfp}, often using physics-based priors derived from partial differential equations (PDE). Leveraging the connections between the SDF and equations such as Eikonal PDE and the Poisson PDE~\citep{vizzo2021poisson, zhong2023shine}, recent works have regularized neural reconstruction through PDE based objectives. While Eikonal and Poisson constraints provide metric consistency to the learned field, noisy and sparsely sampled LiDAR observations can still introduce spatial jitter in the reconstructed surfaces.

A fundamentally different perspective on surface characterization arises from smoothness priors derived from energy functionals. Neural reconstruction approaches have used Hessians and their energies as penalties \cite{zhang2022critical, calakli2011ssd} but often lose geometric detail due to excessive smoothing \cite{wang2023aligning}. This is addressed by instead imposing constraints on the singular Hessian \cite{wang2023neural} or by aligning gradients and the Hessians \cite{wang2023aligning}. However, these approaches struggle in sparse and large-scale scenes. A key challenge in LiDAR sweeps is that we often have large open regions spanning meters, sampled sparsely and noisily. In these settings, Eikonal constraints can capture larger geometric structures, but noise manifests as surface-level jitter. Meanwhile, classical geometry processing has long leveraged optimization strategies over surface curvature energies defined by shape operators \cite{crane2013robust, desbrun1999implicit, gruber2020computational, elsey2009analogue, stein2020smoothness}, which enable surface smoothing, fairing, bending, and the tuning of surface-level irregularities \cite{soliman2026constant, neunteufel2023numerical}, but their use in large-scale LiDAR mapping remains unexplored.

\textbf{Our Contribution.} 
Motivated by these classical techniques, in this work, we pose neural LiDAR mapping as a surface energy optimization problem defined over the parameter space of neural representations. We consider the energy functional defined over the shape operator, which measures the surface variations. We find that this energy can characterize the sharp, jitter seen on the reconstructed surface and optimizing over it seeks a minimum energy configuration for reconstruction. By applying this energy as the regularizer for a sparse hierarchical neural implicit representation, we show that the resulting maps exhibit improved geometric fidelity across large-scale datasets. We evaluate our approach on a diverse set of object-, building-, and street-scale scenes using established LiDAR mapping benchmarks, including KITTI \cite{geiger2013vision}, MaiCity \cite{vizzo2021icra}, and Newer College \cite{ramezani2020newer}. To evaluate reconstruction of more complex surfaces, we additionally construct a synthetic LiDAR benchmark from the Tanks-and-Temples \cite{knapitsch2017tanks} dataset. We show that our method consistently achieves the highest overall F-scores across these benchmarks, outperforming both LiDAR mapping methods and neural surface reconstruction approaches.
\section{Background}
\label{sec:formatting}

\subsection{Standard and Neural LiDAR mapping}
Typical LiDAR mapping approaches relies on a variety of geometric and probabilistic techniques. These involve correspondence point optimization \cite{zhang2014loam, vizzo2023kiss}, incremental triangulation \cite{lin2023immesh}, Gaussian Process-based meshing \cite{jia2024cad, ruan2023slamesh}, surfel-based meshing \cite{behley2018efficient}, and coverage maps \cite{stachniss2003mapping}. Additionally, volumetric Truncated Signed Distance Function (TSDF) integration frameworks, such as VDBFusion \cite{vizzo2022vdbfusion}, provide efficient spatial data structures for merging range sensor data. While effective, these explicit methods often struggle to produce globally smooth manifolds from highly sparse data.

To address the limitations of discrete meshing, implicit neural representations have emerged as powerful continuous surrogates for SDF estimation. Neural Radiance Fields (NeRFs) and Neural LiDAR Fields (NFL) \cite{huang2023nfl} model scenes through coordinate-based networks \cite{azinovic2022neural}. For memory-efficient large-scale offline mapping, methods such as SHINE \cite{zhong2023shine}, 3QFP \cite{sun20243qfp}, PIN-SLAM \cite{pan2024pin}, and LONER \cite{isaacson2023loner} introduce sparse, tree-based data structures to accelerate neural implicit reconstruction. 
\subsection{Physics Based Approaches}
 The emerging physics-based mapping techniques integrate mathematical priors into neural networks to constrain the underlying geometry of the learned manifold. Among these, data-driven approaches \cite{huang2022neural} use gradient-based data-dependent kernels within the deep learning framework \citep{huang2023neural, williams2021neural, williams2022neural}, while physics-informed approaches embed the governing differential equations like the Eikonal \cite{wang2024neurodin, zhong2023shine} or Poisson constraints \cite{park2023p} as loss functions to guide learning. PUMA \cite{vizzo2021poisson} leverages Poisson constraints for LiDAR mapping. Fully data-driven extensions utilize neural operators  \cite{kovachki2023neural, azizzadenesheli2024neural, viswanath2023neural} to implicitly model the underlying PDEs across various graphics tasks \cite{andrade2023neural, viswanath2024reduced, chetan2025accurate, bangaru2022differentiable}.

\subsection{Neural Signed Distance Functions}
Neural SDFs for surface reconstruction often leverage higher-order gradient terms for improving geometric fidelity. The volumetric Hessian energy, introduced for surface reconstruction by \cite{calakli2011ssd}, is applied to neural fields by RegSDF \cite{zhang2022critical}, which penalizes the Hessian of the implicit function directly, minimizing the Hessian energy together with a minimal-surface term. \cite{wang2023aligning} penalizes the product of the Hessian and the local gradient, which vanishes when the gradient is an eigenvector of the Hessian. Neural-Singular-Hessian \cite{wang2023neural} minimizes the determinant of the Hessian near the surface. But these approaches can fail in large-scale and highly sparse settings. Other methods include penalizing the divergence of the gradient field \cite{ben2022digs}, leveraging periodic activations \cite{sitzmann2020implicit}, symmetric Chamfer data terms \cite{harenstam2024diffcd}, perturbed query coordinates \cite{ling2025stochastic}, and neural Galerkin solvers \cite{xi2025neuralssd}.

However, it has been shown that directly optimizing Hessian energies can often lead to oversmoothing and eventually mesh degeneracy \cite{wang2023aligning}. In sparse LiDAR settings, ensuring smooth reconstructions while avoiding mesh degeneracy remains a key challenge. We propose to leverage the shape operators \cite{crane2013robust, droske2004level, lim2012novel, novello2022exploring, yoshizawa2002fair}, wherein, we define the variational energy over the curvature of the level set. We detail this in \cref{sec:problem_setup}.

\section{Problem Setup}
\label{sec:problem_setup}
Let $\chi = \{x_i \in \mathbb{R}^3, i=1,...,N\}$ be the point cloud sampled from the spatial domain $\Omega$, where $\Omega \subset \mathbb{R}^3$ denotes a bounded spatial domain equipped with the standard Euclidean metric.  We seek to extract a surface mesh $\mathbf{M} = (\mathbf{V}, \mathbf{F})$, with vertices $\mathbf{V}$ and faces $\mathbf{F}$, such that $\mathbf{M}$ accurately reconstructs the underlying geometry of the scene. We define this surface as the zero-level set of a continuous SDF, $u: \mathbb{R}^3 \rightarrow \mathbb{R}$, denoted by $\mathcal{S} = \{x \in \Omega \mid u(x) = 0\}$. The discrete mesh $\mathbf{M}$ is extracted by applying marching cubes to the signed distance field $u$. Our objective is to learn the implicit mapping $\mathcal{G}_\theta: \chi \rightarrow u$, where $\mathcal{G}$ is a neural network parameterized by learnable weights $\theta$.

\subsection{Preliminaries}

We consider scalar fields $u : \Omega \rightarrow \mathbb{R}$, where $u$ represents the SDF parameterized by a neural implicit representation. Throughout this work, we assume $u \in H^2(\Omega)$, the Sobolev space of square-integrable functions whose weak derivatives up to second order are also square integrable. The Hessian matrix of $u$ is defined as
$[\mathcal{H}(u)]_{ij} = \partial^2 u/\partial x_i \partial x_j$.

The shape operator $S$, as used in graphics \cite{crane2013robust, droske2004level, lim2012novel, novello2022exploring}, defines the curvature of a surface. For the field $u$, it is expressed through the unit normal $n = \nabla u / \|\nabla u\|$ and the tangent projector $P = I - n n^{\top}$ as
\begin{equation}
S = \frac{P\,\mathcal{H}(u)\,P}{\|\nabla u\|},
\label{eq:shape_operator}
\end{equation}
whose eigenvalues are the principal curvatures $\kappa_1$, $\kappa_2$ and zero, so that $\|S\|_F^2 = \kappa_1^2 + \kappa_2^2$ is the total curvature of the level set \cite{crane2013robust, yoshizawa2002fair}. This can be expressed as a variational energy over the surface $\Sigma$,
\begin{equation}
E_{S}(u) = \int_{\Sigma} \|S\|_F \, dA,
\label{eq:surface_energy}
\end{equation}
whose minimizers define a smooth surface \cite{yoshizawa2002fair, corman2024curvature, rowe2024sharpening}.

\subsection{Neural Energy Optimization}

In this section, we formalize the smoothness energy optimization for large-scale LiDAR settings. Let $\mathcal G_\theta$ denote the neural implicit representation of the signed distance field $u$, parameterized by learnable weights $\theta$, so that the network parameterizes the admissible function space. Let $\mathcal{X}_{pool} = \{(x_i, s_i)\}_{i=1}^M$ denote the finite set of samples drawn along the sensor rays together with their SDF labels. At each optimization step we draw a uniform random subset $\mathcal{B} \subset \mathcal{X}_{pool}$ of size $N$, and estimate the surface energy over $\mathcal{B}$. Because $\mathcal G_\theta$ is differentiable with respect to its input coordinates, $\nabla u$ and $\mathcal{H}(u)$ are evaluated directly by automatic differentiation.

\vspace{2pt}\noindent\textbf{Surface energy Optimization}
We estimate \cref{eq:surface_energy} by its sample mean over the batch, $E_{S}(\theta) = |\mathcal{B}|^{-1}\sum_{x \in \mathcal{B}} \|S(x)\|_F$, with $S$ as in \cref{eq:shape_operator}. When used alongside the reconstruction losses, the optimizer serves as the minimizer of this expression and imposes smoothness constraints.  

\vspace{2pt}\noindent\textbf{Reconstruction constraints}
The SDF is typically modeled as the solution to the Eikonal equation and is the standard physics-informed loss used in surface reconstruction. Similarly, as LiDAR surfaces tend to be non-watertight, a common loss used to ensure the interpolated points remain compact is the minimal surface area constraint used in \cite{zhang2022critical}. This term helps compact geometry interpolated between sparse observations: $E_{M}(\theta) = |\mathcal{B}|^{-1}\sum_{x \in \mathcal{B}} (\varepsilon/\pi)\,/\,(\varepsilon^{2} + u(x)^{2})$ where $u(x)$ is the predicted signed distance at the sample $x$, and $\varepsilon = 0.08\,\mathrm{m}$ is the width of the band
around the surface over which area is charged. We refer the reader to
\cite{zhang2022critical} for the derivation of this term.


\begin{figure*}[ht!]\centering\includegraphics[width=1\linewidth]{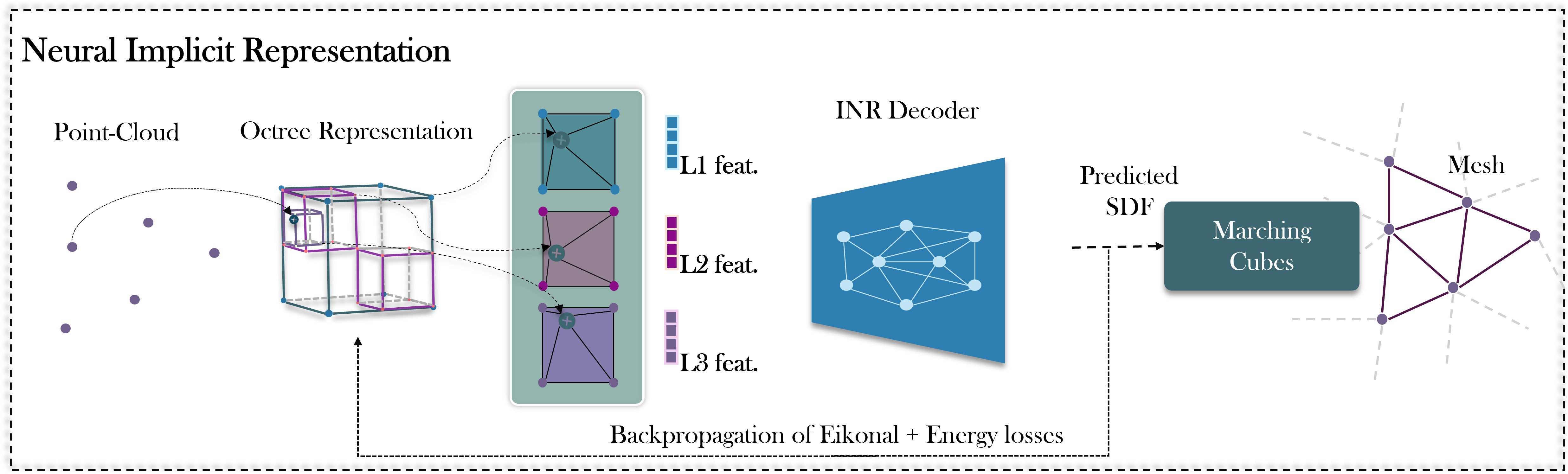}\caption{\textbf{Framework architecture:} Our architecture consists of an implicit neural representation and a learnable octree, trained using the continuous smoothness energy as a geometric supervision signal.}\label{fig:architecture}\end{figure*}

\begin{figure*}[ht!]
    \centering
    \includegraphics[width=1\linewidth]{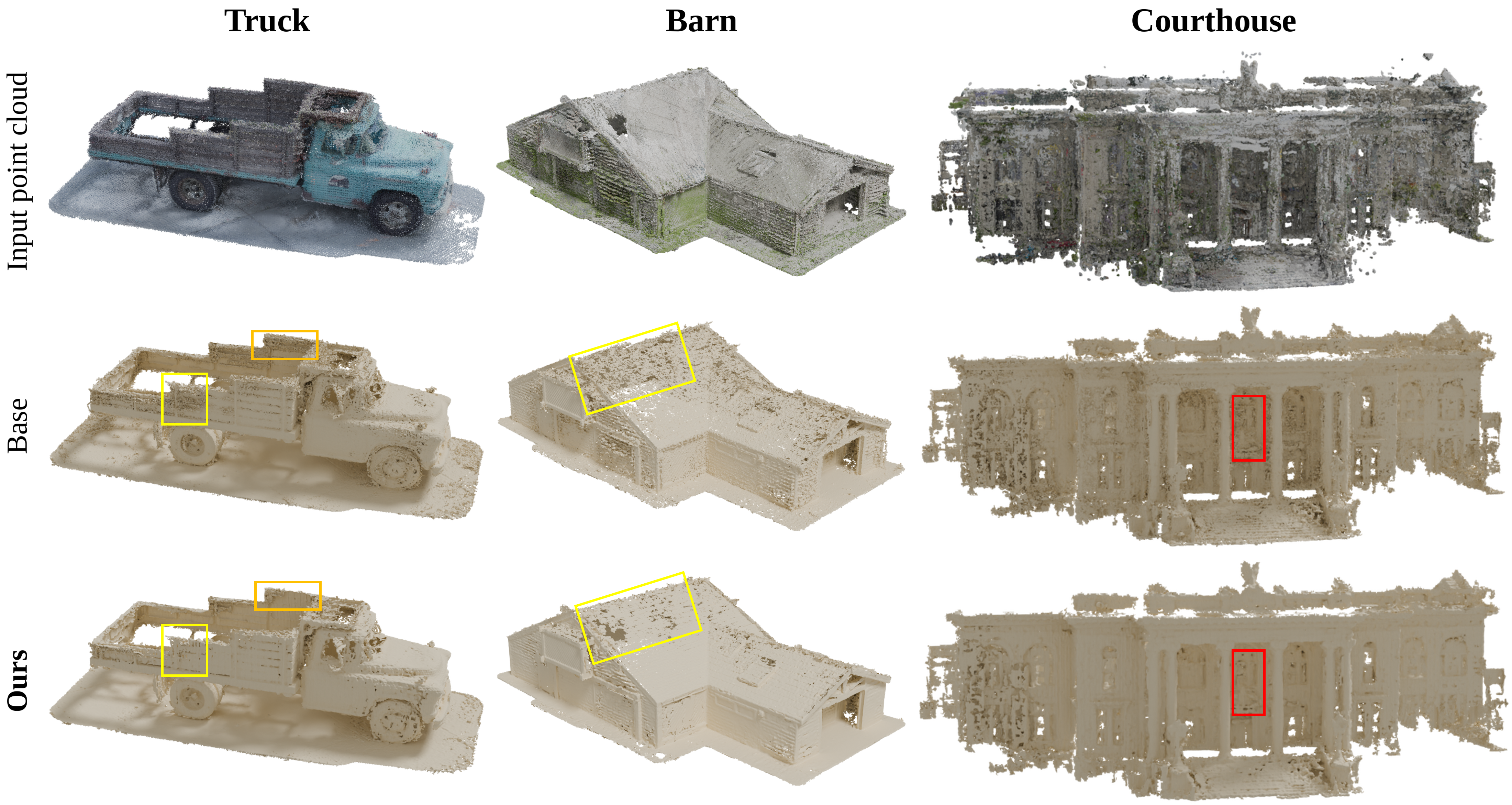}
    \caption{\textbf{Qualitative comparison across scenes.} One column per scene (\textit{Truck}, \textit{Barn}, \textit{Courthouse}): input cloud, \emph{Base}, and ours. All three share a sampler, octree and decoder, so only the regularizer differs.}
    \label{fig:qualitative_res}
\end{figure*}
\section{Training and Implementation Details}



\paragraph {Dataset Details}

We evaluate our approach on standard scene-level LiDAR sweeps. The MaiCity dataset \cite{vizzo2021icra} is a synthetic dataset generated from a 3D CAD model of an urban environment. It contains 100 frames, which were created by simulating LiDAR scans through ray-casting techniques. The Newer College dataset \cite{ramezani2020newer} consists of data captured at New College, Oxford, using a handheld sensor setup, which includes a stereo camera and a 3D LiDAR. This dataset includes 1,500 frames and is accompanied by precise ground truth measurements. Finally, we also benchmark against the KITTI dataset \cite{geiger2013vision}, a widely used dataset for outdoor odometry tasks. KITTI includes sequences captured with a 64-beam LiDAR sensor, along with camera data, and provides near ground truth estimates for incremental mapping in both urban and rural environments.

Existing LiDAR sets are often captured at street-level or at building-level. We aim to understand the impact of surface energy optimization on more complex structures. We therefore consider Tanks-and-Temples dataset \cite{knapitsch2017tanks}, whose training scenes are captured with outside-looking-in trajectories and span sculptures, large vehicles and house-scale
buildings, and whose per-scene evaluation thresholds run from $3$\,mm to $25$\,mm, a range. This set contains a survey-grade laser scan per scene together with an
evaluation volume and an alignment matrix. We use five of its scenes: \textit{Ignatius},
\textit{Truck}, \textit{Caterpillar}, \textit{Barn} and \textit{Courthouse}, representng the natural progression from object level to scene level captures. We additionally consider the indoor \textit{Redwood} scene \cite{choi2015robust}.

\paragraph{Constructing a synthetic LiDAR set from Tanks-and-Temples}
For each scene, Tanks-and-Temples provides an image set, a camera trajectory, and a laser
scan with an evaluation volume. The image set is the input a method is expected to
reconstruct from, and the laser scan is held out as the evaluation reference. To construct a
synthetic LiDAR set, we convert the image set into a per-frame point cloud. We use COLMAP \cite{fisher2021colmap} to
reconstruct each scene from the published camera poses, running feature extraction,
sequential matching and point triangulation, followed by dense multi-view stereo. This
produces a depth map per image, which we back-project into a per-frame point cloud paired
with that frame's camera centre. Each frame is cropped to the evaluation volume published
with the scene. The transform published with each scene includes a scale factor, which we
apply to the points, so that distances in the converted scene are in metres. Camera
intrinsics are estimated per scene: we project the ground-truth points into each image at a
candidate focal length, take the outline of the projected region, and score the mean image
gradient along it, choosing the focal length with the highest score.

\paragraph{Feature Encoding and Sampling:} Across these datasets, we adopt the hierarchical parameterization strategy established by recent offline implicit models \cite{zhong2023shine, sun20243qfp}. We parameterize the SDF using a differentiable hierarchical octree-based feature grid. Multi-resolution features are stored in hash tables indexed by Morton codes, enabling $\mathcal{O}(1)$ query access without dense memory allocation. For a given 3D point $x_i$, the octree features are passed into a shared 2-layer MLP with smooth activations in between the layers. For ground-truth supervision, we compute the ray-aligned signed distance $d_i$ from the sensor origin to the measured LiDAR point cloud. Following the sampling strategy established in SHINE \cite{zhong2023shine}, we recognize that precise SDF values far from the surface have very little impact on the final geometry, whereas values close to zero define the surfaces and should exert a higher impact. Thus, the distance is mapped to an occupancy probability $l_i \in [0, 1]$ using a sigmoid function with a truncation hyperparameter $\sigma$, allowing us to densely sample training points within a $\pm 3\sigma$ band around the surface. For the octree and the architecture parameterizations we use the hyperparameters and training setups specified in SHINE \cite{zhong2023shine}, which we provide in Appendix section C. 

Our framework is designed for unbounded, incremental large-scale mapping. Consequently, the computational domain $\Omega$ dynamically expands as the sensor traverses the environment. Because the map is unbounded, we do not impose classical Dirichlet or Neumann constraints on a global boundary $\partial \Omega$. Instead, the energy prior is evaluated point-wise over the aggregated coordinate pool (managed via our sliding-window sampler). This ensures the smoothness constraint is enforced locally.

\paragraph{Loss Formulation.} Our architecture, based on SHINE \cite{zhong2023shine}, is optimized end-to-end using a joint objective. We use the Binary Cross-Entropy (BCE) formulation of SHINE, which applies a sigmoid mapping to the predicted signed distance values, $(o_i=\sigma(\mathcal G_\theta(x_i)))$. This formulation reflects the observation that surface reconstruction is governed primarily by the zero-level set, while errors far from the surface have little influence on the geometry. 
\begin{equation}
    \mathcal{L}_{bce} = \frac{1}{N} \sum_{i=1}^N \left[ l_i \log(o_i) + (1 - l_i) \log(1 - o_i) \right]
\end{equation}
where, $l_i$ denotes the label computed by applying sigmoid to the projected distance to the surface $l_i = \sigma(d_i)$. In addition to the above loss, the standard Eikonal loss from SHINE is employed to enforce the metric property of the signed distance field through its spatial gradients. These two terms combine with the terms of \cref{sec:problem_setup} to give the objective of \cref{eq:total_loss}. In practice, second-order derivatives exhibit larger numerical magnitudes than first-order losses (\cref{tab:hessian_ablation}), motivating the use of smaller weights. 

\vspace{2pt}\noindent\textbf{Total objective.}
The three constraints are combined with the binary cross-entropy data term from \cite{zhong2023shine} and the overall loss is given by 
\begin{equation}
\mathcal{L} = \lambda_1 \mathcal{L}_{bce} + \lambda_2 \mathcal{L}_{Eikonal}
            + \lambda_{S} E_{S} + \lambda_{M} E_{M},
\label{eq:total_loss}
\end{equation}
and the variational problem becomes the finite-dimensional optimization $\theta^\ast = \arg\min_\theta \mathcal{L}(\theta)$. $\lambda_{S}$ and $\lambda_{M}$ are set per scene so that each term contributes a fixed fraction of the total loss at initialization, measured on the untrained field. We provide the theoretical background in Appendix \cref{sec:appendix_theory} and the optimization algorithm in Appendix \cref{sec:appendix_algorithm}. Following \cite{zhong2023shine}, we set $\lambda_1 = 0.1$ and $\lambda_2 = 0.1$. The single-term ablation of \cref{tab:hessian_ablation} studies the sensitivity of that choice.

Code and configurations for all reported experiments will be released.
\begin{figure*}[h!]
    \centering
    \includegraphics[width=1\linewidth]{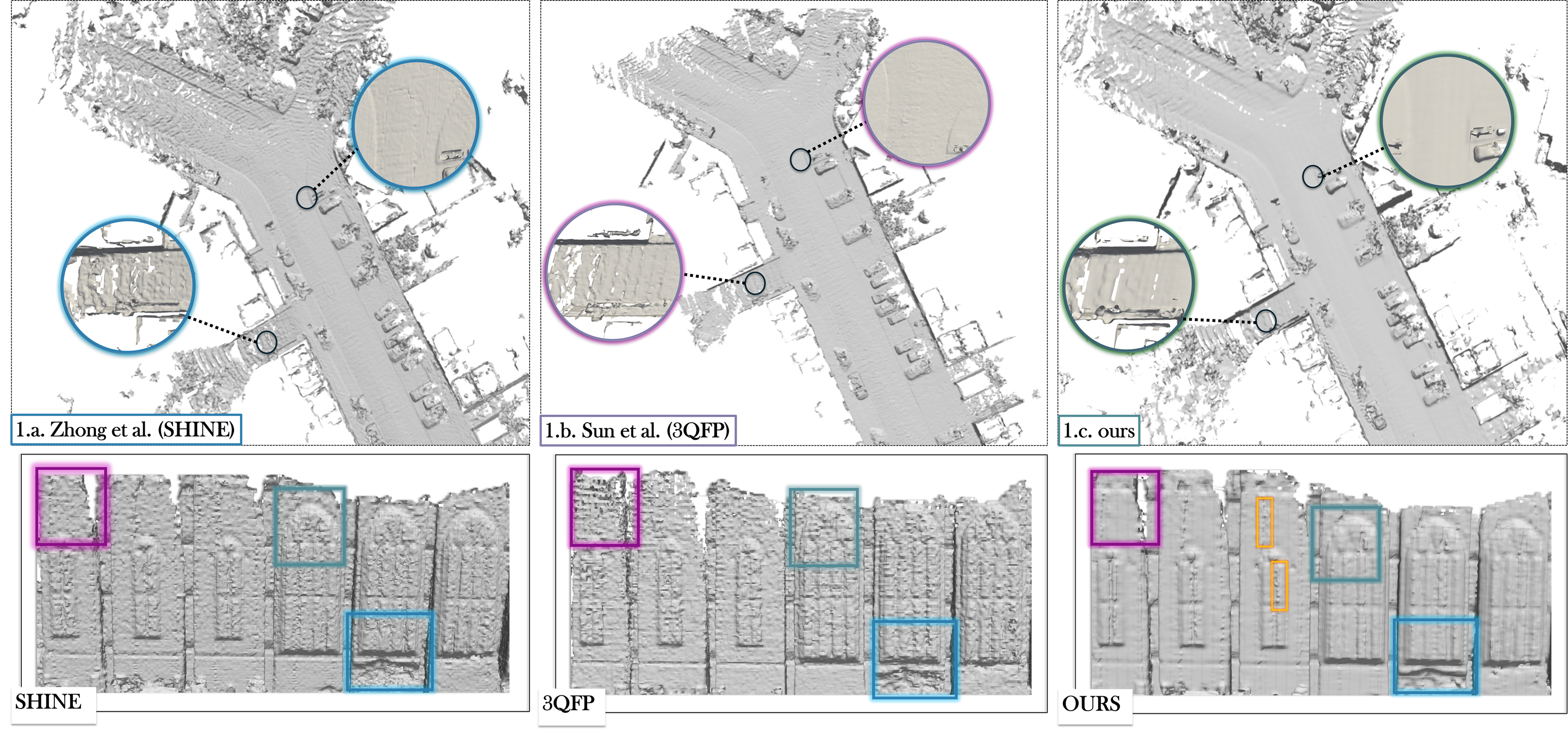}
    \caption{\textbf{Neural implicit reconstruction via energy optimization.} (Top) KITTI 00, bird's-eye. (Bottom) Newer College, front view.}
    \label{fig:Introfig}
\end{figure*}

\paragraph{Evaluation protocol:}
After training, the mesh is extracted by querying the field on a regular grid spanning the
occupied octree nodes and running marching cubes, so the vertices span points beyond the
training set. Accuracy is measured from the mesh vertices to the laser scan, and
completeness from the scan to the mesh surface using exact point-to-triangle distance, both
restricted to the published evaluation volume.

\paragraph{Evaluation Metrics:} To quantitatively assess the fidelity of our neural implicit reconstructions, we leverage the standard reconstruction metrics from the prior works \cite{zhong2023shine}. Accuracy (Acc.) measures the median distance from the reconstructed surface to the ground-truth point cloud, whereas Completeness (Comp.) quantifies the median distance from the ground-truth points to the reconstructed surface, both evaluated against the surface itself using a bounding volume hierarchy. The Chamfer-$L_1$ (CD-$L_1$) distance provides a comprehensive bidirectional nearest-neighbor error. We compute Precision (Acc. Ratio) and Recall (Comp. Ratio) at a per-scene threshold $\tau$ to derive the overall F1 score. Following standard evaluation protocols, we employ an error threshold of 10\,cm for the MaiCity and KITTI datasets, and 20\,cm for the NewerCollege dataset. For KITTI, we utilize an aggregated projection of the point-cloud frames as the ground truth, whereas NewerCollege and MaiCity provide precomputed ground-truth point clouds.

\section{Experiments}


\begin{table}[ht!]
\vspace{5pt}
\caption{Quantitative results on \textit{NewerCollege}, \textit{KITTI} and \textit{MaiCity}. Threshold $10$\,cm (MaiCity, KITTI) and $20$\,cm (NewerCollege). Bold is best.}
\label{tab:main_res}
\resizebox{8.7cm}{!}{
\begin{tabular}{clcccccc}
\toprule
\multicolumn{1}{l}{\textbf{DATA}} & \textbf{MODEL} & \textbf{\begin{tabular}[c]{@{}c@{}}ACC.\\ {[}cm{]}$\downarrow$\end{tabular}} & \textbf{\begin{tabular}[c]{@{}c@{}}COMP.\\ {[}cm{]}$\downarrow$\end{tabular}} & \textbf{\begin{tabular}[c]{@{}c@{}}C-L1\\ {[}m{]}$\downarrow$\end{tabular}} & \textbf{\begin{tabular}[c]{@{}c@{}}ACC. \\ RATIO\\ {[}\%{]}$\uparrow$\end{tabular}} & \textbf{\begin{tabular}[c]{@{}c@{}}COMP. \\ RATIO\\ {[}\%{]}$\uparrow$\end{tabular}} & \textbf{\begin{tabular}[c]{@{}c@{}}F\\ SCORE\\ {[}\%{]}$\uparrow$\end{tabular}} \\ \midrule
\multirow{7}{*}{\textit{\begin{tabular}[c]{@{}c@{}}Newer\\ College\end{tabular}}} & PUMA & 7.70 & 15.37 & 0.107 & 89.77 & 80.31 & 84.77 \\
 & \begin{tabular}[c]{@{}l@{}}VDB\\ FUSION\end{tabular} & 7.10 & 12.40 & 0.095 & 92.81 & 88.56 & 90.63 \\
 & NKSR & 20.10 & 33.70 & 0.168 & 54.77 & 44.38 & 49.03 \\
 & \begin{tabular}[c]{@{}l@{}}PIN\\ SLAM\end{tabular} & 11.04 & \textbf{9.30} & 0.137 & 62.92 & \textbf{96.67} & 76.22 \\
 & 3QFP & 6.72 & 10.50 & 0.097 & 93.63 & 93.35 & 93.48 \\
 & SHINE & 6.70 & 10.00 & 0.084 & 93.80 & 93.60 & 93.70 \\
  \rowcolor{softgreen} 
 & \textbf{OURS} & \textbf{5.68} & 11.18 & \textbf{0.082} & \textbf{97.08} & 91.88 & \textbf{94.41} \\ \midrule
\multirow{5}{*}{\textit{Kitti}} & ImMesh & 8.65 & 18.67 & 0.077 & 70.20 & 60.33 & 64.89 \\
 & \begin{tabular}[c]{@{}l@{}}PIN\\ SLAM\end{tabular} & 12.33 & 25.81 & 0.09 & 66.43 & 56.70 & 61.18 \\
 & 3QFP & 7.24 & 4.63 & 0.059 & 70.48 & 92.25 & 79.91 \\
 & SHINE & 7.04 & 4.62 & 0.058 & 72.18 & 92.70 & 81.16 \\
 \rowcolor{softgreen} 
 & \textbf{OURS} & \textbf{6.23} & \textbf{4.39} & \textbf{0.056} & \textbf{75.74} & \textbf{93.96} & \textbf{83.87} \\ \midrule
\multirow{7}{*}{\textit{MaiCity}} & ImMesh & 6.78 & 6.10 & 0.077 & 73.06 & 84.42 & 78.33 \\
 & PUMA & 3.50 & 27.58 & 0.056 & 93.01 & 57.45 & 71.02 \\
 & \begin{tabular}[c]{@{}l@{}}VDB\\ FUSION\end{tabular} & 4.00 & 7.10 & 0.044 & 93.01 & 78.22 & 84.97 \\
 & \begin{tabular}[c]{@{}l@{}}PIN\\ SLAM\end{tabular} & 7.39 & 5.50 & 0.0812 & 61.36 & 88.75 & 72.56 \\
 & 3QFP & 1.51 & 3.20 & 0.031 & 94.91 & \textbf{95.61} & 95.25 \\
 & SHINE & 1.11 & \textbf{3.25} & 0.029 & 96.61 & 95.20 & 95.90 \\
  \rowcolor{softgreen} 
 & \textbf{OURS} & \textbf{1.07} & 3.79 & \textbf{0.024} & \textbf{97.94} & 94.17 & \textbf{96.01} \\ \bottomrule
\end{tabular}}
\end{table}

\subsection{Quantitative Results}

\begin{table}[h!]
\centering
\caption{Ablation study evaluating the impact of different physics-informed formulations on the \textit{Newer College} dataset.}
\label{tab:regularization_ablation}
\resizebox{\columnwidth}{!}{
\begin{tabular}{l | ccccc}
\toprule
\textbf{Method} & \textbf{\begin{tabular}[c]{@{}c@{}}ACC.\\ {[}cm{]}$\downarrow$\end{tabular}} & \textbf{\begin{tabular}[c]{@{}c@{}}COMP.\\ {[}cm{]}$\downarrow$\end{tabular}} & \textbf{\begin{tabular}[c]{@{}c@{}}C-L1\\ {[}m{]}$\downarrow$\end{tabular}} & \textbf{\begin{tabular}[c]{@{}c@{}}ACC. \\ RATIO\\ {[}\%{]}$\uparrow$\end{tabular}} & \textbf{\begin{tabular}[c]{@{}c@{}}F\\ SCORE\\ {[}\%{]}$\uparrow$\end{tabular}} \\
\midrule
Eikonal & 6.70 & 10.00 & 0.084 & 93.80 & 93.70 \\
Poisson & 6.55 & 18.25 & 0.141 & 95.63 & 63.09 \\
Energy Only & 6.68 & 17.12 & 0.125 & 93.05 & 65.27 \\
Eikonal + Dirichlet Energy & 6.57 & 11.41 & 0.083 & 95.12 & 93.39 \\
\midrule
\rowcolor{softgreen} \textbf{Eikonal + Surface Energy} & \textbf{5.68} & 11.18 & \textbf{0.082} & \textbf{97.08 }& \textbf{94.41} \\
\bottomrule
\end{tabular}
}
\end{table}
\subsubsection{LiDAR mapping baselines}
We begin by discussing how our proposed approach compares against LiDAR mapping baselines.  While baseline architectures approximate a valid SDF, they often lack a governing prior to manage the structural noise inherent in massive, noisy point clouds. As reported in \cref{tab:main_res} (where \textbf{ours} refers to energy optimized SHINE framework), our framework improves surface Accuracy and F-scores across all three benchmarks. Compared to the strongest performing baselines, our approach yields accuracy improvements of approximately 11.5\% on \textit{KITTI} (from 7.04 cm to 6.23 cm) and 15.2\% on \textit{Newer College Dataset} (from 6.70 cm to 5.68 cm). When processing millions of LiDAR points, where prior models overfit to sparse regions and produce topological jitter, the energy formulation biases the network toward well-conditioned, smooth SDF. In contrast, we find \textit{MaiCity} to have more marginal improvements. We observe that this is because the baselines already perform well on this synthetic dataset, where the points are largely well structured. We provide qualitative visual results in \cref{fig:Introfig} as well as Appendix \cref{fig:qualitative_ncd_maicity}. We find that our proposed approach improves over regularizations defined by Poisson forms \cite{park2023p}, and Dirichlet energy forms \cite{stein2020simple}, as shown in \cref{tab:regularization_ablation}. First, \emph{Energy Only} optimization performs poorly because it lacks the Eikonal constraint, which enforces the metric structure of the signed distance field and anchors the implicit representation to a valid distance function. Second, the energy operator is sensitive to the high-frequency fluctuations present in the observations, producing gradient magnitudes several orders larger than the primary reconstruction losses. Thus, the energy term must be heavily down-weighted, with the optimum occurring near ($3e^{-9}$). In this regime, it operates not as a competing metric constraint, but as a principled geometric inductive bias. It resolves the ill-posedness of sparse point cloud mapping by identifying the geometrically smoothest and most structurally stable manifold from the family of valid distance fields satisfying the Eikonal constraint. Lastly, simpler first-order or Laplacian-based regularizers provide weaker geometric constraints, which is consistent with prior works \cite{stein2018natural, stein2020smoothness}. 
\subsubsection{Neural Reconstruction Baselines}
To analyze our approach more rigorously, we compare our optimization approach with eight baselines - RegSDF \cite{zhang2022critical}, DiGS \cite{ben2022digs}, SIREN \cite{sitzmann2020implicit}, GradHess \cite{wang2023aligning}, Singular Hessian \cite{wang2023neural}, DiffCD \cite{harenstam2024diffcd}, Stochastic Preconditioning \cite{ling2025stochastic} and NeuralSSD \cite{huang2022neural}. To ensure fairness in comparison, we setup all these approaches over the base SHINE architecture (Denoted by Base).  We leverage the Tanks-and-Temples dataset, which provides objects of increasing complexity, ranging from vehicles to building complexes. This enables us to analyze how the approach behaves in objects with complex curved surfaces such as trucks, and on large planar surfaces seen in buildings. Besides F-score, Chamfer distance and normal-augmented F-score from \cite{zhang2022critical}, we measure the vertex count and, the energy reduction ($\Delta E$), which is the ratio of curvature energy of the regularization method to the baseline SHINE method. Full definitions are provided in Appendix \cref{sec:appendix_table_conventions}. We present the results in \cref{tab:results_main}, with visual results in \cref{fig:Introfig}, \cref{fig:qualitative_res} with additional visuals in the Appendix. 

Our key observations are that our approach consistently exhibits the highest F1-score as well as normal-aligned F1-score while reducing the Chamfer distance. We observe that  RegSDF and NeuralSSD generate surfaces which exhibit the highest reduction of energy as well as the lowest vertex count. This is consistent with the known oversmoothing issues observed in these methods \cite{wang2023neural, wang2023aligning}. We observe that this results in the formation of holes within the oversmoothed surface, as shown in Appendix \cref{fig:ablation_grid}. In contrast, DiffCD exhibits moderate energy reduction, with improved surface variance ratio, but struggles to resolve the jitter completely, as shown in \cref{fig:energy_row}. Similarly, SIREN, Singular Hessian and stochastic preconditioning exhibit moderate to minimal changes. 
\providecommand{\tbd}{\textcolor{gray}{--}}
\providecommand{\nax}{\textcolor{gray}{---}}
\providecommand{\tba}{\textcolor{gray}{\textsc{tba}}}

\begin{table}[t]
\centering
\scriptsize
\caption{\textbf{Reconstruction metrics and field characteristics.} Quantitative comparisons against neural reconstruction baselines. $^\dagger$ denotes settings with lower vertex count and lower energies, but higher Chamfer distance and lower F-scores than baseline, indicating oversmoothing.}
\label{tab:results_main}
\setlength{\tabcolsep}{5.1pt}
\begin{tabular}{llrrrrr}
\toprule
\multirow{2}{*}{\textbf{Scene}} &
\multirow{2}{*}{\textbf{Method}} &
\multirow{2}{*}{\shortstack{\textbf{F}\\\textbf{Score$\uparrow$}}} &
\multirow{2}{*}{\shortstack{\textbf{CD-$L_1$}\\(mm)$\downarrow$}} &
\multirow{2}{*}{\shortstack{\textbf{F}\\\textbf{Score$_n$$\uparrow$}}} &
\multirow{2}{*}{\textbf{$\Delta E$}} &
\multirow{2}{*}{\shortstack{\textbf{Vert.}\\\textbf{(M)}}} \\
& & & & & & \\
\midrule

\multirow{10}{*}{\rotatebox[origin=c]{90}{\textbf{Truck}}} 
& Base & 52.4 & 8.64 & 45.8 & \nax & 6.42 \\
& RegSDF~\cite{zhang2022critical} & 51.6 & 8.95 & 49.4 & 2.48$^\dagger$ & 3.40$^\dagger$ \\
& NeuralSSD~\cite{xi2025neuralssd} & 51.7 & 8.79 & 46.3 & 2.88$^\dagger$ & 2.00$^\dagger$ \\
& GradHess~\cite{wang2023aligning} & 52.0 & 8.71 & 46.0 & 1.07 & 5.21 \\
& Sing. Hessian~\cite{wang2023neural} & 52.5 & 8.65 & 45.9 & 0.56 & 5.63 \\
& DiGS~\cite{ben2022digs} & 52.3 & 8.65 & 45.2 & 0.81 & 6.04 \\
& SIREN~\cite{sitzmann2020implicit} & 51.3 & 8.79 & 45.3 & 0.06 & 5.39 \\
& DiffCD~\cite{harenstam2024diffcd} & 52.1 & 8.66 & 45.3 & 1.01 & 6.13 \\
& Stoch. Prec.~\cite{ling2025stochastic} & 52.4 & 8.65 & 45.8 & 1.05 & 5.68 \\
\rowcolor{softgreen}
& \textbf{Ours} & \textbf{54.1} & \textbf{8.59} & \textbf{50.7} & \textbf{1.96} & \textbf{4.39} \\

\midrule

\multirow{10}{*}{\rotatebox[origin=c]{90}{\textbf{Barn}}} 
& Base & 40.8 & 19.55 & 26.6 & \nax & 6.71 \\
& RegSDF~\cite{zhang2022critical} & 37.2 & 20.59 & 26.3 & 2.01$^\dagger$ & 4.37$^\dagger$ \\
& NeuralSSD~\cite{xi2025neuralssd} & 39.9 & 21.07 & 33.7 & 1.58$^\dagger$ & 4.50$^\dagger$ \\
& GradHess~\cite{wang2023aligning} & 40.3 & 19.32 & 29.4 & 0.56 & 9.57 \\
& Sing. Hessian~\cite{wang2023neural} & 42.0 & 19.38 & 34.5 & 0.39 & 9.63 \\
& DiGS~\cite{ben2022digs} & 40.5 & 19.47 & 32.3 & 0.16 & 9.09 \\
& SIREN~\cite{sitzmann2020implicit} & 40.2 & 19.29 & 29.2 & 0.02 & 6.32 \\
& DiffCD~\cite{harenstam2024diffcd} & 40.9 & 19.22 & 26.5 & 0.96 & 7.45 \\
& Stoch. Prec.~\cite{ling2025stochastic} & 41.4 & 19.52 & 34.2 & 0.69 & 10.34 \\
\rowcolor{softgreen}
& \textbf{Ours} & \textbf{43.3} & \textbf{19.19} & \textbf{36.8} & \textbf{1.21 }& \textbf{5.23} \\

\midrule

\multirow{10}{*}{\rotatebox[origin=c]{90}{\textbf{Courthouse}}} 
& Base & 26.0 & 71.09 & 15.5 & \nax & 5.19 \\
& RegSDF~\cite{zhang2022critical} & 24.5 & 71.21 & 15.1 & 1.74$^\dagger$ & 3.42$^\dagger$ \\
& NeuralSSD~\cite{xi2025neuralssd} & 25.9 & 74.14 & 17.2 & 2.08$^\dagger$ & 3.66$^\dagger$ \\
& GradHess~\cite{wang2023aligning} & 26.1 & 72.36 & 16.1 & 1.42 & 4.71 \\
& Sing. Hessian~\cite{wang2023neural} & 26.9 & 71.26 & 15.7 & 0.66 & 5.11 \\
& DiGS~\cite{ben2022digs} & 27.8 & 71.34 & 15.5 & 0.25 & 4.98 \\
& SIREN~\cite{sitzmann2020implicit} & 25.2 & 71.11 & 16.8 & 0.83 & 5.02 \\
& DiffCD~\cite{harenstam2024diffcd} & 27.9 & 70.94 & 15.8 & 0.97 & 5.14 \\
& Stoch. Prec.~\cite{ling2025stochastic} & 27.5 & 71.42 & 15.4 & 1.12 & 4.88 \\
\rowcolor{softgreen}
& \textbf{Ours} & \textbf{28.3} & \textbf{70.32} & \textbf{17.4} & \textbf{1.60} & \textbf{3.69} \\

\bottomrule
\end{tabular}
\end{table}

\textit{Limitation:} While the energy-based approaches resolve high-frequency structural noise, they inherently create a conservative geometric bias regarding surface completeness. Because the objective explicitly penalizes spatial variations in the normal field, which mathematically correspond to large Hessian magnitudes, the network resists high-curvature topologies. Consequently, baselines can achieve marginally higher absolute completeness on specific datasets (e.g., PIN-SLAM reaching 9.30 cm on \textit{NewerCollege} compared to our 11.18 cm, or SHINE achieving 3.25 cm on \textit{MaiCity} vs. our 3.79 cm). Empirically, we observe that this shift toward structurally constrained, noise-reduced mapping directly translates to improved performance and stability in downstream tasks.


\begin{figure*}[t]
    \centering
    \includegraphics[height=3.9cm]{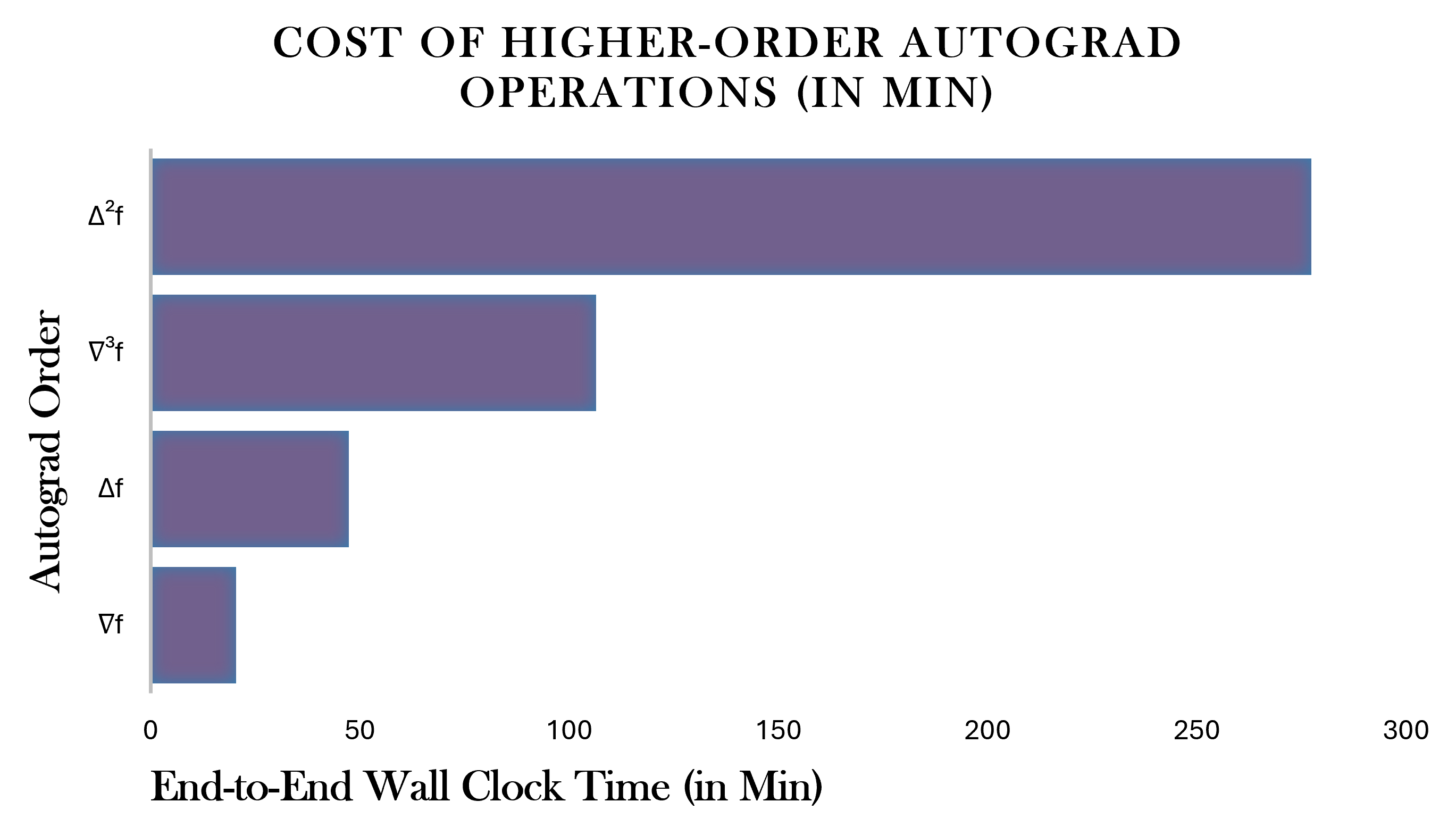}\hfill
    \includegraphics[height=3.9cm]{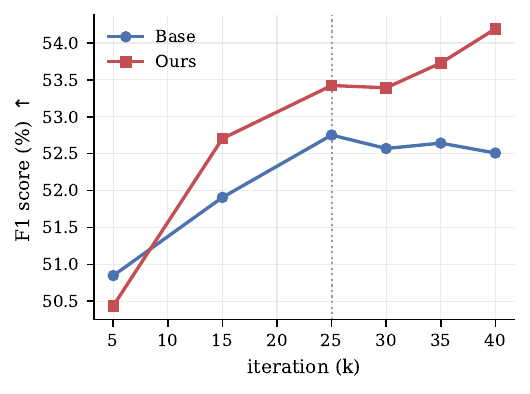}\hfill
    \includegraphics[height=3.9cm]{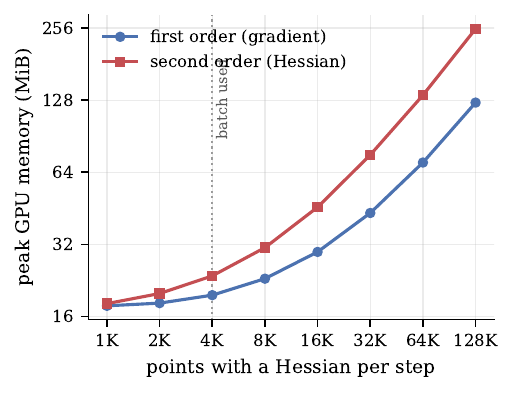}
    \caption{\textbf{Cost and performance of the energy.} \textit{(Left)} Wall-clock for 40K steps against differentiation order. \textit{(Middle)} Accuracy against step on \textit{Truck}. \textit{(Right)} Peak GPU memory against Hessian point count. }
    \label{fig:cost}
\end{figure*}

\subsection{Cost-Performance Tradeoff}
\label{sec:cost_performance}

Embedding higher-order geometric priors inevitably increases the cost of \emph{training}, since the curvature energy requires evaluating second-order spatial derivatives through automatic differentiation. Compared to first-order objectives such as the Eikonal loss, this incurs additional memory consumption and computation during optimization. We investigate this overhead in \cref{fig:cost}, which computes the wall-clock time by simulating \texttt{autograd} operations over 40K steps. The second-order formulation requires approximately 45 minutes of training time compared to 25 minutes for a first-order baseline, reflecting the cost of enforcing second-order geometric regularity.

Importantly, this overhead is incurred \emph{only during training}. The Hessian energy shapes the parameters of the neural implicit representation, embedding the smoothness prior directly into the learned field. Once optimization is complete, inference consists only of evaluating the neural implicit function itself, requiring no Hessian evaluations, higher-order derivatives, or auxiliary optimization procedures. Consequently, mesh extraction and downstream deployment incur no additional computational overhead beyond that of a conventional neural implicit representation.

As an academic point of comparison, one may also consider optimizing the fourth-order biharmonic residual associated with the Euler--Lagrange equation of the surface Hessian energy (also shown in \cref{fig:cost}). Such a formulation requires nested higher-order automatic differentiation and rapidly becomes computationally prohibitive. Our experiments indicate that a direct fourth-order implementation requires nearly 300 minutes of training, yielding approximately a $6.6\times$ increase in runtime relative to our Hessian energy formulation.

What that additional training time buys is shown in the middle panel of \cref{fig:cost}, which evaluates accuracy at intermediate checkpoints, not only at convergence. The two methods separate within the first few thousand iterations: the energy leads by $0.85$ points at $15$k, $0.9$ at $25$k, and $1.5$ at $40$k. The unregularized baseline peaks near $25$k and then drifts back, while the regularized field continues to improve. The prior therefore shapes the optimization trajectory from the outset and does not act as a late correction to an already-converged solution. 

The number of points at which the Hessian is evaluated is the sampler batch size, fixed by the optimizer, so it is independent of scene extent and of the number of captured points. \Cref{fig:cost} (right) compares the peak GPU memory with the number of input points. The optimized SHINE framework does not incur significant GPU overheads for the second-order derivative compared to the first. We contrast memory consumption for different scenes in the Appendix \cref{tab:compute_cost}.

\begin{table}[h!]
\centering
\caption{SLAM accuracy (ATE RMSE, m) against LiDAR odometry (above) and LiDAR SLAM (below) on Newer College. Ours is an energy-optimized PIN-SLAM variant.}
\label{tab:slam_newer_college}
\resizebox{0.5\textwidth}{!}{
\begin{tabular}{c | ccccccc | c}
\toprule
Method & 01 & 02 & quad e & math e & ug e & cloister & stairs & Avg. \\
\midrule
F-LOAM \cite{wang2021f} & 6.74 & - & 0.40 & 0.26 & \underline{0.09} & 7.69 & - & 3.04 \\
KISS-ICP \cite{vizzo2023kiss} & \underline{0.62} & 1.88 & \underline{0.10} & \textbf{0.07} & 0.33 & 0.30 & - & \underline{0.55} \\
\midrule
SuMa \cite{behley2018efficient} & 2.03 & 3.65 & 0.28 & 0.16 & \underline{0.09} & 0.20 & 1.85 & 1.18 \\
MULLS \cite{pan2021mulls} & 2.51 & 8.39 & 0.12 & 0.35 & 0.86 & 0.41 & - & 2.11 \\
MD-SLAM \cite{di2022md} & - & 1.74 & 0.25 & - & - & 0.36 & 0.34 & - \\
SC-LeGO-LOAM \cite{kim2018scan} & - & \underline{1.30} & \textbf{0.09} & - & - & 0.20 & 3.20 & - \\
\midrule
PIN-LO \cite{pan2024pin} & 2.08 & 5.32 & \textbf{0.09} & 0.09 & \textbf{0.07} & \underline{0.19} & \underline{0.07} & 1.13 \\
PIN-SLAM \cite{pan2024pin} & 0.43 & \textbf{0.31} & \textbf{0.09} & 0.09 & \textbf{0.07} & \textbf{0.15} & 0.06 & 0.17 \\
\midrule
\rowcolor{softgreen} \textbf{OURS} & \textbf{0.39} & \textbf{0.31} & \textbf{0.08} & \underline{0.08} & 0.08 & \textbf{0.15} & \textbf{0.04} & \textbf{0.15} \\
\bottomrule
\multicolumn{9}{l}{\footnotesize All sequences are with loops. The best result is bold and the second best is underscored. 7 denotes failure.} \\
\multicolumn{9}{l}{\footnotesize - indicates the number is not reported and unavailable using the open-source code.} \\
\end{tabular}
}
\end{table}

\subsection{Downstream Tasks}

To investigate whether the geometric regularity induced by our energy formulation extends beyond surface reconstruction, we evaluate its impact on Simultaneous Localization and Mapping (SLAM). In neural implicit SLAM systems, pose estimation is performed by directly registering incoming scans against the learned signed distance field (SDF), with the optimization relying on both the predicted distance values and their spatial gradients. Consequently, the quality of the implicit representation can directly influence downstream localization performance. While the Eikonal objective constrains the magnitude of the SDF gradient, our Hessian energy additionally regularizes its spatial variation, promoting a smoother and more coherent implicit geometry throughout the mapped environment.

To quantitatively evaluate this effect, we measure the root-mean-square error (RMSE) of the absolute trajectory error (ATE) using Umeyama alignment on the Newer College benchmark. As shown in \cref{tab:slam_newer_college}, incorporating our energy-optimized implicit representation into the PIN-SLAM mapping backend yields consistent improvements in localization accuracy without modifying the odometry or optimization pipeline. Averaged across all evaluated sequences, our method achieves an ATE RMSE of 0.15\,m compared to 0.17\,m for the original PIN-SLAM formulation, corresponding to an 11.8\% reduction in average trajectory error. Improvements are also observed on several individual sequences, including sequence 01 (0.43\,m $\rightarrow$ 0.39\,m, 9.3\%), \emph{quad e} (0.09\,m $\rightarrow$ 0.08\,m, 11.1\%), and \emph{stairs} (0.06\,m $\rightarrow$ 0.04\,m, 33.3\%). Although the proposed energy formulation is designed as a reconstruction prior and not a localization objective, these results indicate that improving the regularity of the learned implicit field can positively influence downstream registration accuracy, highlighting the utility of energy-optimized neural representations for robotics applications, although a detailed analysis of this interaction remains an interesting direction for future work.

\label{sec:autograd_tradeoff}

\section{Conclusion}

In this work, we introduced an energy-based framework for neural implicit surface mapping that embeds classical smoothness priors directly into continuous signed distance fields. By treating the neural implicit representation as a parameterization of the minimum-energy configuration of a continuous energy functional, our formulation unifies variational geometry processing with modern neural field representations while remaining computationally practical for large-scale LiDAR mapping. Experiments demonstrate that this energy-driven perspective consistently improves reconstruction fidelity and geometric regularity while generalizing naturally to downstream robotics tasks such as localization and mapping. Broadly, we believe this work establishes energy optimization as a principled foundation for integrating classical geometric priors into neural implicit representations, opening new directions for physically grounded 3D scene understanding.

\section*{Acknowledgement}
This material is based upon work supported in part by the DEVCOM Army Research Laboratory under cooperative agreement W911NF2020221.
{
    \small
    \bibliographystyle{ieeenat_fullname}
    \bibliography{main}
}
\clearpage

\newpage
\appendix

\section{Theoretical Background}
\label{sec:appendix_theory}

This section collects the definitions used in \cref{sec:problem_setup}. All are established results, restated here in our notation with their sources.

\paragraph{Shape operator of a level set.}
For $u$ with $\nabla u \neq 0$, the unit normal of the level set is $n = \nabla u / \|\nabla u\|$, and the Jacobian of this normal field is $Dn = P\,\mathcal{H}(u) / \|\nabla u\|$ with $P = I - n n^{\top}$. Restricting $Dn$ to the tangent space gives the shape operator \cite[Eq.~4]{droske2004level},
$$S = \frac{P\,\mathcal{H}(u)\,P}{\|\nabla u\|},$$
whose eigenvalues are the principal curvatures $\kappa_1$, $\kappa_2$ and zero. The same expression is used for neural implicit surfaces in \cite[Eq.~1]{novello2022exploring}.

\paragraph{Bending energy.}
As $S$ is symmetric with eigenvalues $(\kappa_1, \kappa_2, 0)$, its squared Frobenius norm is $\|S\|_F^2 = \kappa_1^2 + \kappa_2^2 = 4H^2 - 2K$, where $H$ and $K$ are the mean and Gaussian curvature of the level set \cite{gruber2020computational}. The integral $\int_{\Sigma} \|S\|_F^2 \, dA$ is the bending energy of the surface, equal to the squared norm of its second fundamental form \cite{crane2013robust, gruber2020computational}. Where $\int_{\Sigma} K \, dA$ is fixed, as for a closed surface by Gauss--Bonnet, this energy is equivalent to the Willmore functional $\int_{\Sigma} H^2 \, dA$ \cite{droske2004level, gruber2020computational}.

\paragraph{The $L_1$ form.}
\cite{rowe2024sharpening} study Hessian smoothness energies under the $L_1$ norm and show that their minimizers are piecewise smooth, with non-smoothness concentrated on a sparse set of ridges, where the corresponding $L_2$ energy produces uniformly smooth minimizers. 

\paragraph{Relation to the ambient Hessian.}
Under the Eikonal condition $\|\nabla u\| = 1$, differentiating $\|\nabla u\|^2 = 1$ gives $\mathcal{H}(u)\nabla u = 0$, so $\nabla u$ spans the null space of $\mathcal{H}(u)$ and its remaining eigenvalues are the principal curvatures \cite{wang2023aligning, wang2023neural}. In that case $\|\mathcal{H}(u)\|_F = \|S\|_F$, and the volumetric Hessian energy of \cite{calakli2011ssd, zhang2022critical} coincides with \cref{eq:surface_energy}.

\section{Optimization Algorithm}
\label{sec:appendix_algorithm}

We optimize the neural implicit representation by evaluating the Hessian prior natively within the parameter space. Because our geometry is parameterized by a neural network, the Hessian operator can be evaluated efficiently through double automatic differentiation. The optimization procedure for a given scene is detailed in Algorithm~\ref{alg:pi_sdf_opt}.

\begin{algorithm}[h!]
\caption{Neural Energy Optimization}
\label{alg:pi_sdf_opt}
\begin{algorithmic}[1]
\REQUIRE Pre-computed coordinate pool $\mathcal{X}_{pool}$, Feature Octree $\mathcal{O}$, MLP Decoder $\mathcal{G}_\theta$, Max Iterations $T$, Batch Size $B$.
\STATE \textbf{Initialize} network parameters $\theta$, octree features, and variance parameter $\sigma$.
\FOR{$t = 1$ \TO $T$}
    \STATE Sample random mini-batch of coordinates $\mathcal{X} \subset \mathcal{X}_{pool}$ of size $B$
    \STATE \textbf{Require Gradient:} $\mathcal{X}$.requires\_grad\_(True)
    \STATE Query multi-resolution features: $F \leftarrow \mathcal{O}(\mathcal{X})$
    \STATE Predict SDF values: $\hat{S} \leftarrow \mathcal{G}_\theta(F)$
    
    \STATE \textit{// Compute First and Second-Order Derivatives}
    \STATE $G \leftarrow \nabla_\mathcal{X} \hat{S}$ \quad \textit{(via first-order autograd)}
    \STATE Initialize Hessian matrix $H \in \mathbb{R}^{B \times 3 \times 3}$
    \FOR{$i = 1$ \TO $3$}
        \STATE $H[:, i, :] \leftarrow \nabla_\mathcal{X} G[:, i]$ \quad \textit{(via second-order autograd)}
    \ENDFOR
    
    \STATE \textit{// Compute Optimization Objectives}
    \STATE $\mathcal{L}_{SDF} \leftarrow \text{BinaryCrossEntropy}(\hat{S}, S_{gt}, \sigma)$
    \STATE $\mathcal{L}_{Eikonal} \leftarrow \frac{1}{B} \sum_{x \in \mathcal{S}} (1 - \|G\|_2)^2$ \quad \textit{(If applicable)}
    \STATE $n \leftarrow G / \|G\|_2$, \quad $P \leftarrow I - n n^{\top}$
    \STATE $\mathbf{S} \leftarrow P H P / \|G\|_2$ \quad \textit{(shape operator, \cref{eq:shape_operator})}
    \STATE $\mathcal{L}_{Surface} \leftarrow \frac{1}{B} \sum \|\mathbf{S}\|_F$
    \STATE $\mathcal{L}_{Minimal} \leftarrow \frac{1}{B} \sum \frac{\varepsilon/\pi}{\varepsilon^{2} + \hat{S}^{2}}, \quad \varepsilon = 1/\varepsilon_m$
    
    \STATE \textit{// Total Loss and Backpropagation}
    \STATE $\mathcal{L}_{total} \leftarrow \mathcal{L}_{SDF} + \lambda_e \mathcal{L}_{Eikonal} + \lambda_s \mathcal{L}_{Surface} + \lambda_m \mathcal{L}_{Minimal}$
    \STATE Backpropagate $\mathcal{L}_{total}$
    \STATE Update $\theta$, $\mathcal{O}$, and $\sigma$ via Adam optimizer
\ENDFOR
\end{algorithmic}
\end{algorithm}
\begin{remark}
The octree features are interpolated with the polynomial scheme implemented in the SHINE codebase \cite{zhong2023shine}, $f(t) = 3t^2 - 2t^3$ for spatial queries. Each axis is passed through $f$ independently, and the eight corner weights are the products of the three resulting factors. Because its first derivative vanishes smoothly at the voxel boundaries ($t \in \{0, 1\}$), the interpolated feature field maintains continuity. When composed with the smooth activations of the MLP decoder, the $\nabla^2_{\mathcal{X}} \hat{S}$ terms are piecewise continuous ($f''(t) = 6 - 12t$), and exist almost everywhere (a.e.) in the Lebesgue sense. This enables the energy to be optimized stably via \texttt{autograd}.
\end{remark}

\subsection{Smoothing Behaviour of the Energy}
\label{sec:appendix_smoothing}

\begin{figure}[h!]
    \centering
    \includegraphics[width=1\linewidth]{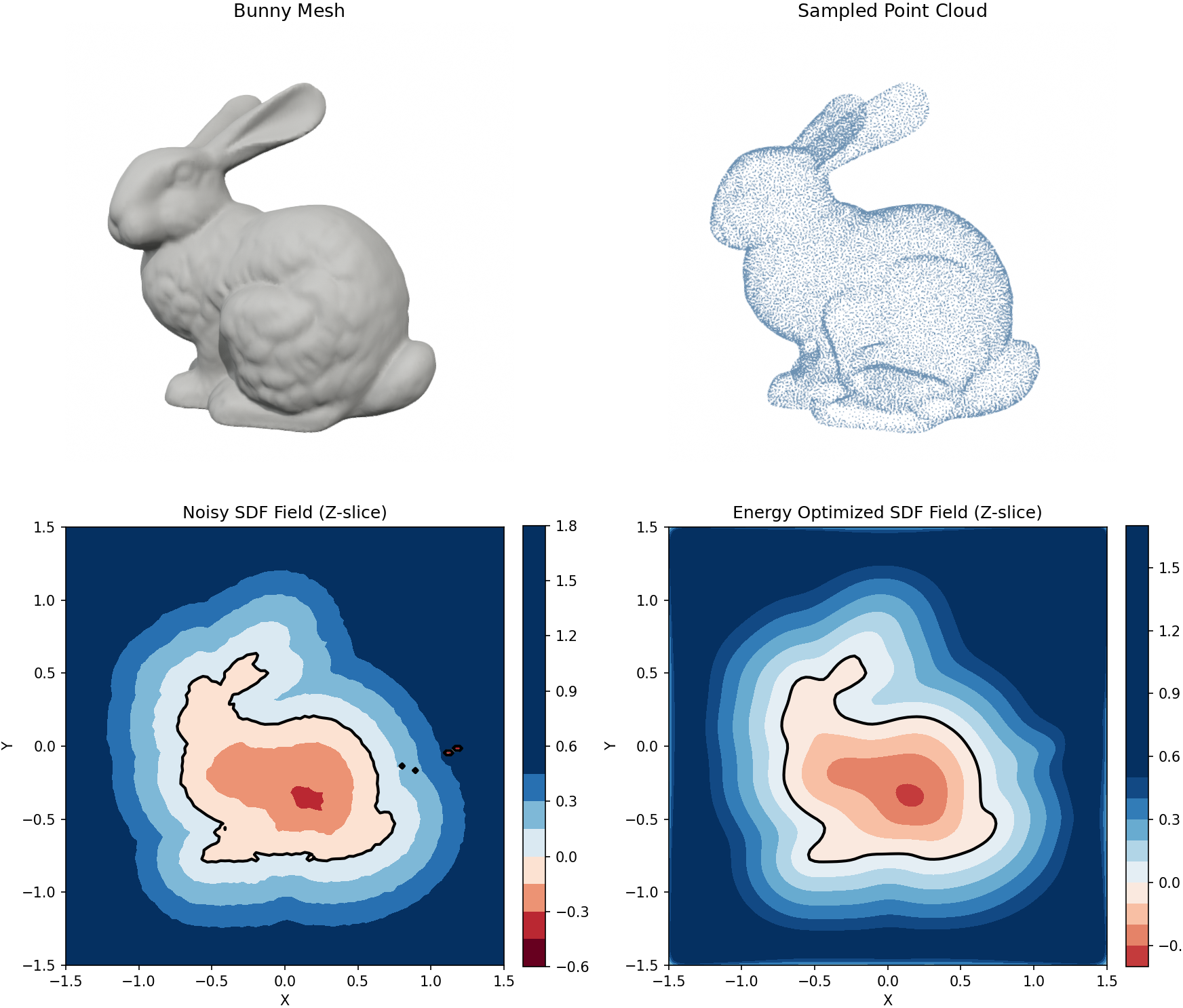}
    \caption{\textbf{Energy optimization in SDF fields.} A noisy field with artifacts
    (left) becomes smooth on minimizing the energy functional (right).}
    \label{fig:sdf_energy_optim}
\end{figure}

The smoothing behaviour induced by the continuous formulation is illustrated in
\cref{fig:sdf_energy_optim}. For a signed distance field corrupted by additive Gaussian
perturbations, direct optimization of the energy removes high-frequency oscillations from
the volumetric field, producing smooth zero-level sets and well-conditioned iso-contours.

\begin{tcolorbox}[title=Connection to Classical Mesh Smoothing,colback=white,colframe=black!60,boxrule=0.5pt]
The proposed formulation is a continuous, mesh-free counterpart to geometric smoothing in
classical mesh processing. Penalizing curvature on a discrete mesh requires
finite-difference operators, lumped mass matrices and conic relaxations
\cite{rowe2024sharpening}. Imposing the prior directly on the continuous signed distance
field evaluates the same quantity through automatic differentiation, so the smoothing
emerges during standard neural optimization without auxiliary variables or specialized
solvers (\cref{fig:sdf_energy_optim}).
\end{tcolorbox}

\begin{table}[h!]
\centering
\small
\caption{\textbf{Cost follows octree occupancy, not scene size.} Feature memory is $C \times d \times 4$ bytes at $d{=}8$.}
\label{tab:compute_cost}
\resizebox{\columnwidth}{!}{
\begin{tabular}{l r r r r}
\toprule
Scene & Finest $\ell$\,(m) & $C$\,(M) & Feat.\,mem\,(MiB) & $\alpha$ \\
\midrule
Ignatius    & 0.005 &  7.41 & 226 & $-2.40$ \\
Truck       & 0.0075 & 16.88 & 515 & $-2.28$ \\
Caterpillar & 0.010 & 16.99 & 518 & $-2.37$ \\
Barn        & 0.020 & 19.71 & 602 & $-2.24$ \\
Courthouse  & 0.050 & 11.72 & 358 & $-2.09$ \\
\bottomrule
\end{tabular}}
\end{table}


\section{Additional Quantitative Results}
\label{sec:appendix_results}

\Cref{tab:results_main} reports the three scenes the qualitative figures accompany:
one object-scale (\textit{Truck}) and two building-scale (\textit{Barn},
\textit{Courthouse}). The remaining scenes are evaluated under an identical
protocol. We consider an additional indoor scene \textit{Redwood}\cite{choi2015robust}, which is RGB-D and its ground truth is registered to the
camera trajectory rather than published in a shared frame, so its absolute distances
carry a larger residual than the Tanks-and-Temples scenes. We present the results in \cref{tab:results_appendix}. 

\providecommand{\tbd}{\textcolor{gray}{--}}
\providecommand{\nax}{\textcolor{gray}{---}}
\providecommand{\tba}{\textcolor{gray}{\textsc{tba}}}

\begin{table*}[t]
\centering
\scriptsize
\caption{\textbf{Reconstruction quality and field regularity, remaining scenes.} Columns and protocol are identical to \cref{tab:results_main}; }
\setlength{\tabcolsep}{10pt}
\begin{tabular}{ll r r r r r}
\toprule
Scene & Method & F-Score\,$\uparrow$ & CD-$L_1$\,$\downarrow$ & F-Score$_n$\,$\uparrow$ & Energy Reduction & Verts\, (M) \\
\midrule
\multirow{10}{*}{\rotatebox[origin=c]{90}{Ignatius}} 
& Base & 67.7 & 3.09 & 59.7 & \nax & 2.03 \\
& RegSDF~\cite{zhang2022critical} & 59.0 & 3.55 & 49.1 & 3.30 & 1.49 \\
& NeuralSSD~\cite{xi2025neuralssd} & 61.6 & 3.45 & 51.4 & 3.56 & 1.29 \\
& GradHess~\cite{wang2023aligning} & 69.1 & 3.05 & 59.7 & 0.97 & 1.78 \\
& Singular Hessian~\cite{wang2023neural} & 68.9 & 3.05 & 61.6 & 0.91 & 1.80 \\
& DiGS~\cite{ben2022digs} & 69.6 & 3.05 & 61.7 & 0.72 & 1.96 \\
& SIREN~\cite{sitzmann2020implicit} & 68.1 & 3.09 & 58.7 & 0.05 & 1.99 \\
& DiffCD~\cite{harenstam2024diffcd} & 70.01 & 3.02 & 62.2 & 1.36 & 2.08 \\
& Stoch. Precond.~\cite{ling2025stochastic} & 69.9 & 3.02 & 62.1 & 1.03 & 1.98 \\
& \textbf{Ours} & \textbf{72.0} & \textbf{3.01} & \textbf{62.3} & 2.70 & 1.64 \\
\midrule
\multirow{10}{*}{\rotatebox[origin=c]{90}{Caterpillar}} 
& Base & 45.1 & 7.89 & 36.4 & \nax & 6.96 \\
& RegSDF~\cite{zhang2022critical} & 43.75 & 8.42 & 37.5 & 2.75 & 3.28 \\
& NeuralSSD~\cite{xi2025neuralssd} & 43.6 & 9.61 & 36.6 & 3.60 & 2.79 \\
& GradHess~\cite{wang2023aligning} & 44.1 & 9.73 & 34.5 & 1.25 & 5.53 \\
& Singular Hessian~\cite{wang2023neural} & 44.3 & 8.05 & 32.0 & 0.57 & 6.55 \\
& DiGS~\cite{ben2022digs} & 45.6 & 7.89 & 31.1 & 0.83 & 6.75 \\
& SIREN~\cite{sitzmann2020implicit} & 45.9 & 8.01 & 32.4 & 0.45 & 6.82 \\
& DiffCD~\cite{harenstam2024diffcd} & 46.3 & \textbf{7.80} & 31.3 & 1.06 & 7.11 \\
& Stoch. Precond.~\cite{ling2025stochastic} & 45.2 & 7.94 & 37.2 & 1.09 & 6.86 \\
& \textbf{Ours} & \textbf{47.9} & 7.91 & \textbf{34.4} & 1.87 & 4.73 \\
\midrule
\multirow{10}{*}{\rotatebox[origin=c]{90}{Redwood}} 
& Base & 54.2 & 34.61 & 45.2 & \nax & 2.92 \\
& RegSDF~\cite{zhang2022critical} & 50.1 & 38.33 & 46.0 & 2.44 & 2.20 \\
& NeuralSSD~\cite{xi2025neuralssd} & 48.3 & 34.52 & 39.0 & 2.49 & 1.82 \\
& GradHess~\cite{wang2023aligning} & 51.2 & 33.84 & 41.5 & 0.06 & 2.75 \\
& Singular Hessian~\cite{wang2023neural} & 55.0 & 34.09 & 44.7 & 0.27 & 3.04 \\
& DiGS~\cite{ben2022digs} & 54.1 & 34.62 & 44.4 & 0.59 & 2.96 \\
& SIREN~\cite{sitzmann2020implicit} & 54.8 & \textbf{33.99} & 45.1 & 0.52 & 2.97 \\
& DiffCD~\cite{harenstam2024diffcd} & 54.3 & 34.46 & 45.2 & 1.39 & 2.93 \\
& Stoch. Precond.~\cite{ling2025stochastic} & 54.1 & 34.67 & 45.1 & 1.00 & 2.92 \\
& \textbf{Ours} & \textbf{56.5} & 34.83 & \textbf{46.7} & 1.69 & 2.63 \\
\bottomrule
\end{tabular}

\label{tab:results_appendix}
\end{table*}

\begin{table}[h!]
\centering
\caption{Ablation of the energy weight on \textit{Newer College}. Scales are set proportional to the nominal Eikonal loss magnitude.}
\label{tab:hessian_ablation}
\resizebox{8.3cm}{!}{
\begin{tabular}{l | c c c c}
\toprule
\textbf{WEIGHT SCALE} & \textbf{$e^{-5}$} & \textbf{$e^{-7}$} & \textbf{$e^{-9}$} & \textbf{$e^{-11}$} \\
\midrule
\textit{Avg. Scaled Energy Magnitude} & 2.86 & 7.8$e^{-1}$ & 9.7e$^{-2}$ & $1.5e^{-3}$ \\
\textit{Reconstruction F Score [\%]} & 5.44 & 64.73 & 93.95 & 91.44 \\
\bottomrule
\end{tabular}
}
\end{table}

\begin{figure}[h!]
    \centering
    \includegraphics[width=1\linewidth]{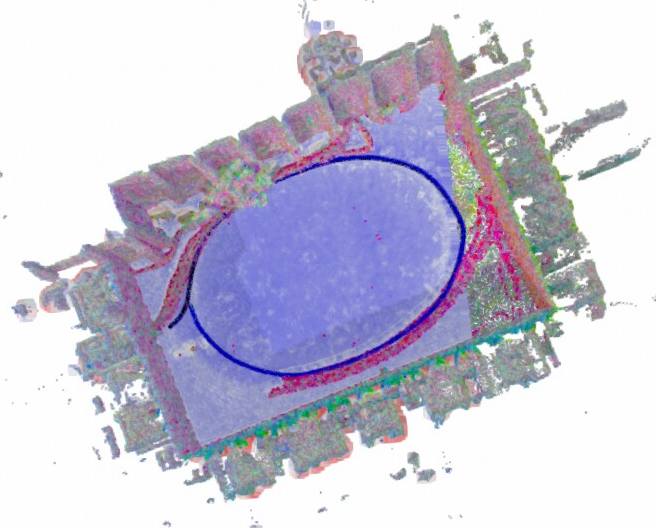}
    \caption{\textbf{Energy-optimized neural points.} Smoothness energy embedded in \texttt{PIN-SLAM} on \textit{Newer College}; estimated trajectory in blue.}
    \label{fig:slam_ncd}
\end{figure}

\subsection{Baseline weight ablation}
\label{sec:appendix_wsweep}

\begin{figure*}[t]
    \centering
    \includegraphics[width=\linewidth]{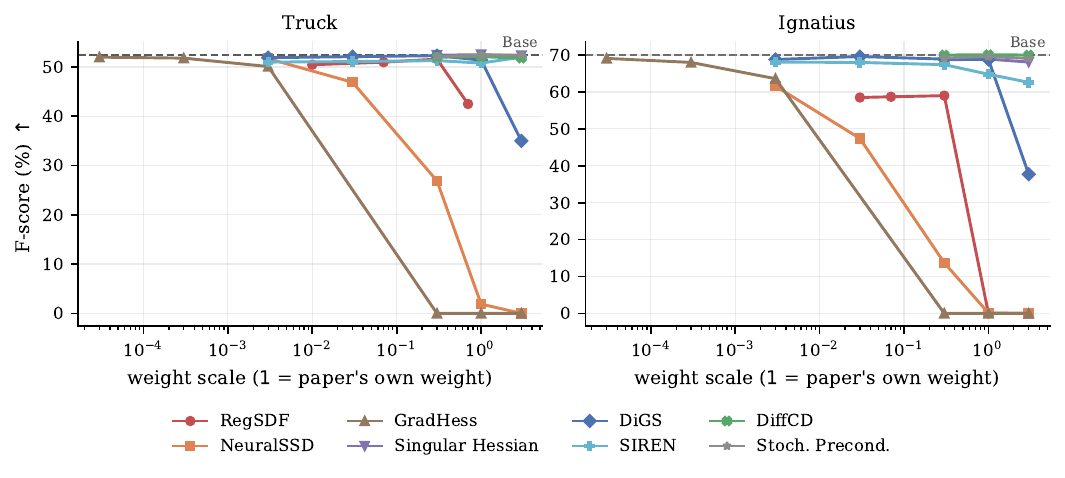}
    \caption{\textbf{Baseline weight profiles.} F-score against the scale applied to each
    regularizer's own declared weight, octree and decoder. A scale of $1$ is the weight published by the corresponding work; the
    declared weights differ by orders of magnitude between methods, and
    \texttt{stoch\_precond\_sigma\_m} is a perturbation radius in metres rather than a loss weight, so the shared axis is relative. }
    \label{fig:ablation_wsweep}
\end{figure*}

Each compared regularizer is swept about its own declared weight, under an identical sampler, octree and decoder
(\cref{fig:ablation_wsweep}). A scale of $1$ is the weight the corresponding
paper reports, so the absolute settings differ by orders of magnitude between
methods and the sweep is plotted on a logarithmic weight axis. We present the optimized settings in \cref{tab:tandt_method_params} with other hyperparameters in \cref{tab:tandt_scene_params}. 

\section{Additional Qualitative Results}
\label{sec:appendix_qualitative}

\begin{figure*}[ht!]
    \centering
    \includegraphics[width=1\linewidth]{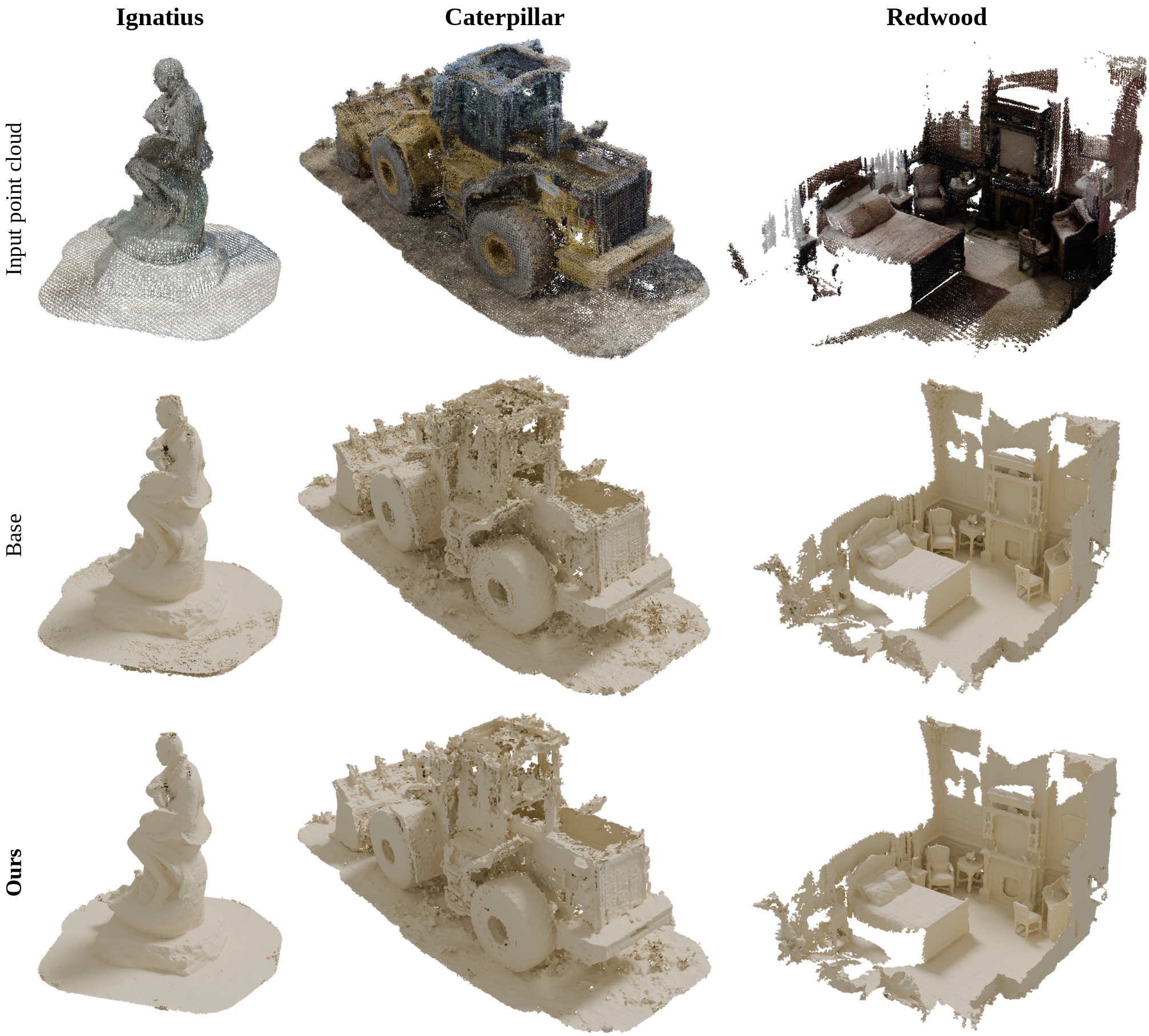}
    \caption{\textbf{Qualitative comparison, remaining scenes.} One column per scene (\textit{Ignatius}, \textit{Caterpillar}, \textit{Redwood}): input point cloud, \emph{Base}, and ours. \textit{Redwood} is indoor RGB-D, and its world frame is the first camera pose, so it is re-oriented to a level floor before framing.}
    \label{fig:qualitative_res_appendix}
\end{figure*}

\Cref{fig:qualitative_res_appendix} repeats the comparison of
\cref{fig:qualitative_res} on the remaining scenes. As there, columns share a
viewpoint per scene and are scaled with the scene's own extent, so apparent size
tracks physical size rather than filling the panel.

\begin{figure*}[ht!]
    \centering
    \includegraphics[width=1\linewidth]{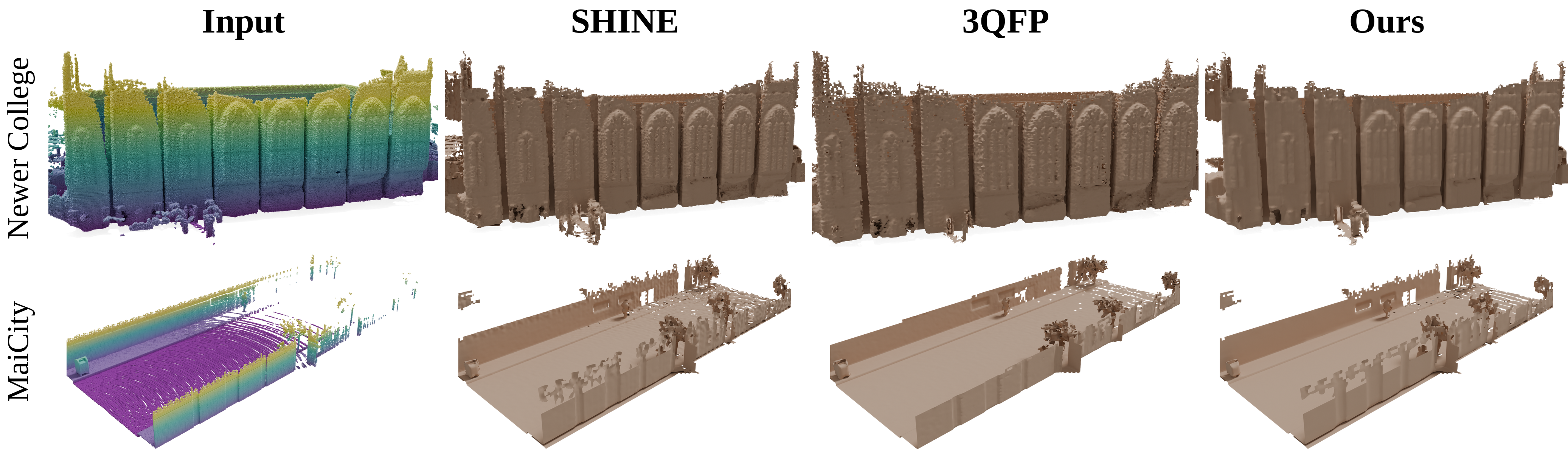}
    \caption{\textbf{Qualitative comparison on the LiDAR benchmarks.} Rows: \textit{Newer College}, \textit{MaiCity}. Columns: input cloud (coloured by height), SHINE~\citep{zhong2023shine}, 3QFP~\citep{sun20243qfp}, ours. SHINE is retrained here from the authors' released configuration, so all three share the input, poses and a $10$\,cm marching-cubes grid.}
    \label{fig:qualitative_ncd_maicity}
\end{figure*}

\Cref{fig:qualitative_ncd_maicity} presents visual comparisons across LiDAR mapping
benchmarks of \cref{tab:main_res}. \textit{Newer College} is shown from outside the quad
looking in, under grazing illumination chosen so that the window reveals cast
shadow; \textit{MaiCity} is a $30\times30$\,m section of the street. Neither
dataset provides colour -- MaiCity is synthetic LiDAR with no intensity channel
and Newer College's fused cloud carries none -- so the input panels are shaded by
height rather than by measured radiometry. We provide close up comparisons in \cref{fig:Introfig}. 

\subsection{Per-scene curvature maps}
\label{sec:appendix_energy_rows}

\begin{figure*}[ht!]
    \centering
    \includegraphics[width=1\linewidth]{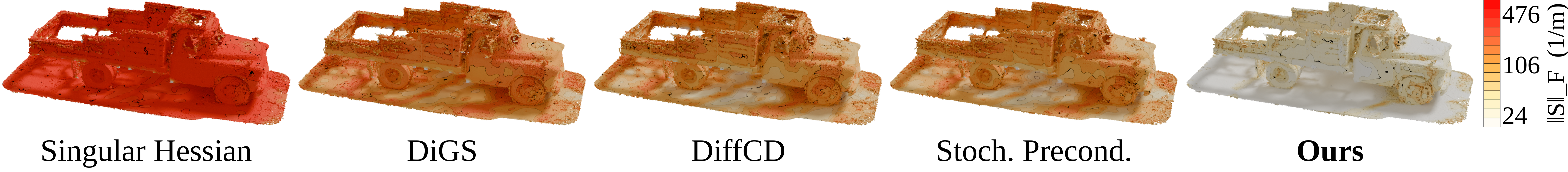}
    \caption{Level-set curvature $\|S\|_F$ on \textit{Truck}. Range $24$--$476$\,m$^{-1}$.}
    \label{fig:energy_row_truck}
\end{figure*}

\begin{figure*}[ht!]
    \centering
    \includegraphics[width=1\linewidth]{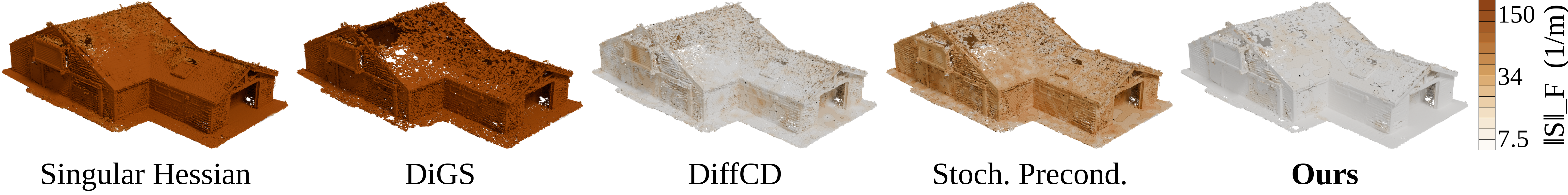}
    \caption{Level-set curvature $\|S\|_F$ on \textit{Barn}. Range $7.5$--$150$\,m$^{-1}$.}
    \label{fig:energy_row_barn}
\end{figure*}

\begin{figure*}[ht!]
    \centering
    \includegraphics[width=1\linewidth]{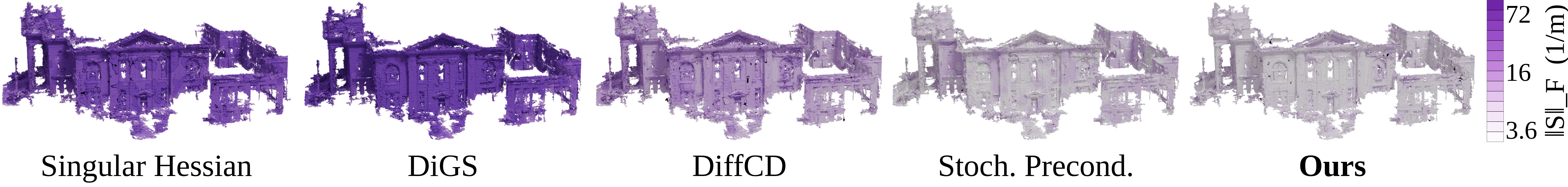}
    \caption{Level-set curvature $\|S\|_F$ on \textit{Courthouse}.}
    \label{fig:energy_row_courthouse}
\end{figure*}

\begin{figure*}[ht!]
    \centering
    \includegraphics[width=1\linewidth]{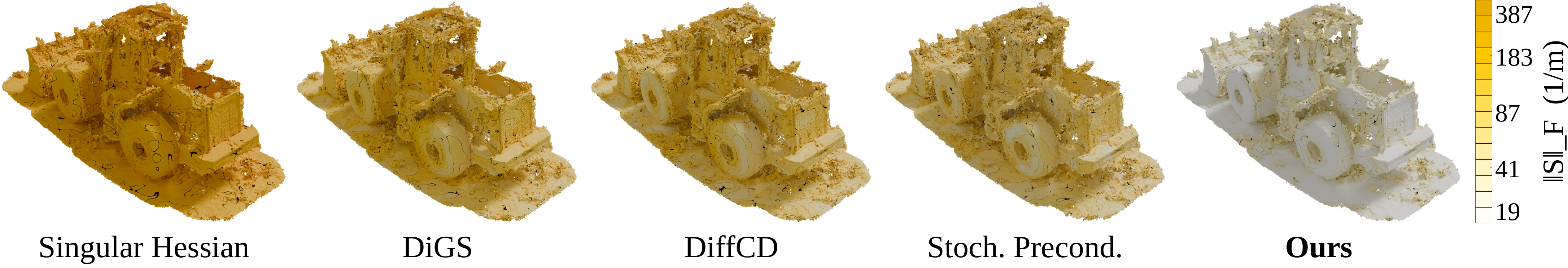}
    \caption{Level-set curvature $\|S\|_F$ on \textit{Caterpillar}.}
    \label{fig:energy_row_caterpillar}
\end{figure*}

\begin{figure*}[ht!]
    \centering
    \includegraphics[width=1\linewidth]{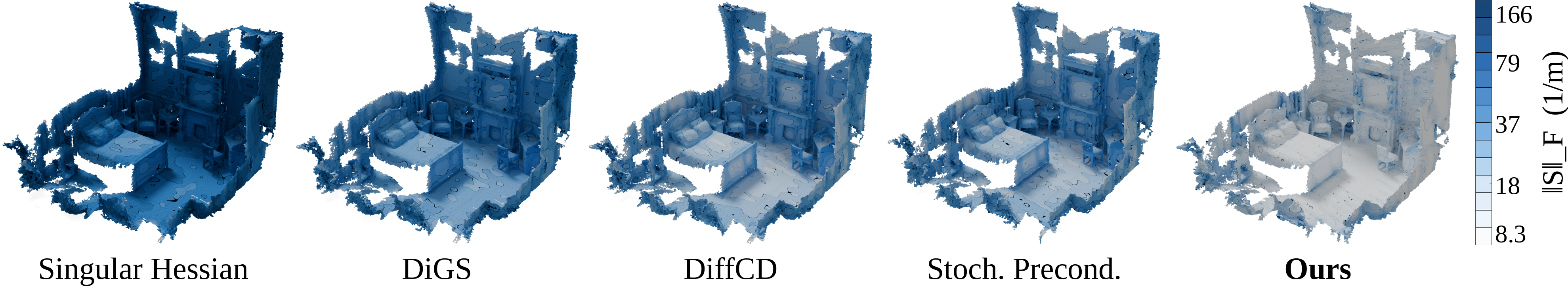}
    \caption{Level-set curvature $\|S\|_F$ on \textit{Redwood}. Absolute values are not
    comparable with the Tanks-and-Temples scenes: this is an indoor RGB-D reconstruction
    at a much coarser threshold.}
    \label{fig:energy_row_redwood}
\end{figure*}

\begin{figure}[h!]
    \centering
    \includegraphics[width=0.8\columnwidth]{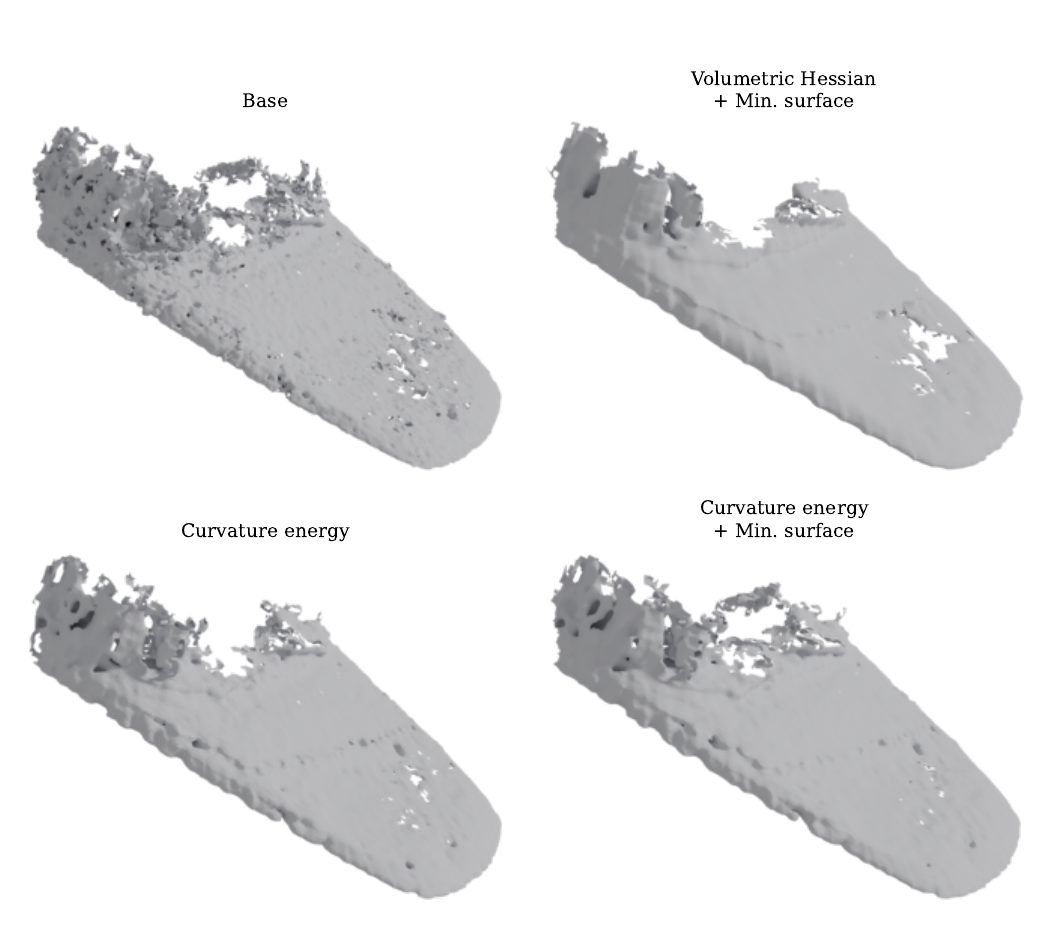}
    \caption{\textbf{Qualitative effect of each energy term.} A close-up of the hood panel of \textit{Truck} under four objectives, all sharing a sampler, octree and
    decoder. The unregularized surface is pitted; directly regularizing via Hessian energy exhibits oversmoothing, resulting in a hole; the curvature energy removes it while the surface stays
    closed.}
    \label{fig:ablation_grid}
\end{figure}

\Cref{fig:energy_row} shows the level-set curvature $\|S\|_F$ on \textit{Ignatius};
the same visualization is presented in 
\cref{fig:energy_row_truck}, \cref{fig:energy_row_barn}, \cref{fig:energy_row_courthouse}, 
\cref{fig:energy_row_caterpillar}, \cref{fig:energy_row_redwood}, for every remaining scene. Within a row a lighter panel is smoother than the unregularized reconstruction
and a darker one is rougher. \textbf{The scale is per scene and the ranges differ by
more than an order of magnitude between them} (for example $24$--$476$\,m$^{-1}$ on
\textit{Truck} against $7.5$--$150$\,m$^{-1}$ on \textit{Barn}), so panels are
comparable within a row and never across rows. Curvature alone does not establish
quality: a method can lower it by reconstructing less surface, which is why these
panels are read together with the other metrics. In \cref{fig:ablation_grid}, we highlight the impact of oversmoothing caused by directly penalizing the Hessian energy, which introduces holes into the reconstructed surface. 

\section{Implementation and Experimental Details}
\label{sec:appendix_exp_details}

To manage memory effectively and ensure scalable optimization, we employ a sliding-window data sampler and a multi-resolution feature octree. The structural parameters of the decoder remain constant across all scenes: a 2-layer MLP with a hidden dimension of 32, querying a feature octree with an 8-dimensional feature vector per node. 

\subsection{Dataset-Specific Hyperparameters}To accommodate the varying spatial scale and sensor densities of different environments, specific hyperparameters were tuned per dataset (summarized in Table~\ref{tab:hyperparams}).
\begin{itemize}
\item \textbf{KITTI (Sequence 00):} Due to the massive scale of the outdoor trajectory, we utilized a slightly larger octree leaf voxel size of $0.3$ m. The network was optimized for $40,000$ iterations with an initial learning rate of $0.05$.
\item \textbf{MaiCity \& Newer College (NCD):} These datasets feature higher local density and tighter structural boundaries. We utilized a finer octree leaf voxel size of $0.2$ m and a reduced learning rate of $0.01$. Optimization converged effectively at $20,000$ iterations.
\item \textbf{Other Scenes: } The other settings were all trained for $40,000$ iterations. 
\end{itemize}

Across all datasets, the energy weights were kept uniform ($\lambda_h = 3e^{-9}, \lambda_e = 0.1$) to demonstrate the robustness and scale-invariance of our continuous formulation. By maintaining a constant Eikonal weight, the network is universally forced to represent a valid metric distance field ($\|\nabla u\|_2 \approx 1$), allowing the Hessian energy to act strictly as a second-order smoothness prior without competing for the fundamental gradient structure. All input point clouds underwent a uniform voxel downsampling with a resolution of $0.05$ m prior to network evaluation.

\begin{table}[h!]
\centering

\resizebox{\columnwidth}{!}{%
\begin{tabular}{lccc}
\toprule
\textbf{Parameter} & \textbf{KITTI} & \textbf{MaiCity} & \textbf{Newer College} \\ \midrule
Octree Leaf Voxel Size & 0.3 m & 0.2 m & 0.2 m \\
Voxel Downsampling & 0.05 m & 0.05 m & 0.05 m \\
Batch Size & 4096 & 4096 & 4096 \\
Iterations & 40,000 & 20,000 & 20,000 \\
Learning Rate & 0.05 & 0.01 & 0.01 \\
SDF Loss Type & BCE & BCE & BCE \\
Energy Loss Weight & $3e^{-9}$ & $3e^{-9}$ & $3e^{-9}$ \\
Eikonal Loss Weight ($\lambda_e$) & 0.1 & 0.1 & 0.1 \\
\bottomrule
\end{tabular}%
}
\caption{Hyperparameter configurations across evaluation datasets.}
\label{tab:hyperparams}
\end{table}

\subsection{Metric Definitions and Table Conventions}
\label{sec:appendix_table_conventions}
The conventions below apply to \cref{tab:results_main} and \cref{tab:results_appendix}.

\textbf{Accuracy and completeness.} Both are exact point-to-surface distances computed
with a BVH, not distances to a point sample of the mesh. A point sample carries a
discretisation floor of roughly $0.5/\sqrt{\text{density}}$, which on these scenes is
comparable to the threshold $\tau$ itself, so it would measure tessellation rather than
geometry. Accuracy is point-to-\emph{point} on the predicted side because several ground
truths are unstructured clouds with no faces; completeness is point-to-surface. The two
are therefore not perfectly symmetric, which we state rather than paper over.

\textbf{Ground-truth point counts (M),} which Comp.\ and Cov.\ Ret.\ are measured against:
Truck 8.07, Barn 12.75, Courthouse 59.14, Ignatius 5.02, Caterpillar 7.26, Redwood 16.89.

\subsection{Complete Parameter Listing (all scenes, all methods)}
\label{sec:appendix_full_params}
\Cref{tab:tandt_scene_params} and \cref{tab:tandt_method_params} give every training and
evaluation parameter used for the six scenes of \cref{tab:results_main} and
\cref{tab:results_appendix}. Both are generated directly from the configuration file that
each run wrote into its own output directory, so they record what was executed rather than
what was intended. 

\providecommand{\tba}{\textcolor{gray}{\textsc{tba}}}
\begin{table*}[t]
\centering
\scriptsize
\setlength{\tabcolsep}{4pt}
\begin{tabular}{lrrrrrr}
\toprule
Parameter & Truck & Ignatius & Caterpillar & Barn & Courthouse & Redwood \\
\midrule
\multicolumn{7}{l}{\textit{Input}} \\
Frame stride (\texttt{every\_frame}) & 5 & 2 & 6 & 3 & 3 & 16 \\
Crop radius $r_{\mathrm{pc}}$ (m) & 8 & 8 & 8 & 15 & 40 & 8 \\
Min.\ range (m) & 0.1 & 0.1 & 0.1 & 0.1 & 0.1 & 0.2 \\
Voxel downsample (m) & 0.008 & 0.005 & 0.01 & 0.01 & 0.025 & 0.01 \\
Global $z$ shift (m) & 7.8 & 7.3 & -157.65 & 21.3 & 17.79 & -- \\
\multicolumn{7}{l}{\textit{Octree}} \\
Leaf voxel size (m) & 0.008 & 0.005 & 0.01 & 0.02 & 0.05 & 0.02 \\
Tree levels (world / feat.) & 12 / 4 & 12 / 4 & 12 / 4 & 13 / 4 & 13 / 4 & 12 / 4 \\
Feature dim. & 8 & 8 & 8 & 8 & 8 & 8 \\
Polynomial interpolation & \checkmark & \checkmark & \checkmark & \checkmark & \checkmark & \checkmark \\
\multicolumn{7}{l}{\textit{Sampler}} \\
Surface sample range (m) & 0.024 & 0.015 & 0.03 & 0.03 & 0.075 & 0.05 \\
Surface samples / ray & 3 & 3 & 3 & 3 & 3 & 3 \\
Free-space begin ratio & 0.5 & 0.5 & 0.5 & 0.5 & 0.5 & 0.5 \\
Free-space end dist.\ (m) & 0.16 & 0.1 & 0.2 & 0.2 & 0.5 & 0.3 \\
Free samples / ray & 3 & 3 & 3 & 3 & 3 & 3 \\
\multicolumn{7}{l}{\textit{Loss (shared)}} \\
Main loss & sdf\_bce & sdf\_bce & sdf\_bce & sdf\_bce & sdf\_bce & sdf\_bce \\
$\sigma_{\mathrm{sigmoid}}$ (m) & 0.008 & 0.005 & 0.01 & 0.01 & 0.025 & 0.02 \\
Eikonal weight $\lambda_e$ & 0.1 & 0.1 & 0.1 & 0.1 & 0.1 & 0.1 \\
\multicolumn{7}{l}{\textit{Optimizer}} \\
Batch size & 4096 & 4096 & 4096 & 4096 & 4096 & 4096 \\
Weight decay & $1\mathrm{e}{-7}$ & $1\mathrm{e}{-7}$ & $1\mathrm{e}{-7}$ & $1\mathrm{e}{-7}$ & $1\mathrm{e}{-7}$ & $1\mathrm{e}{-7}$ \\
\multicolumn{7}{l}{\textit{Marching cubes}} \\
$\mathrm{mc\_res}$ (m) & 0.008 & 0.005 & 0.01 & 0.01 & 0.025 & 0.01 \\
Octree-masked MC & \checkmark & \checkmark & \checkmark & \checkmark & \checkmark & \checkmark \\
MC visibility level & 2 & 2 & 2 & 2 & 2 & 2 \\
\multicolumn{7}{l}{\textit{Evaluation protocol}} \\
Threshold $\tau$ (m) & 0.005 & 0.003 & 0.005 & 0.01 & 0.025 & 0.05 \\
GT voxel size (m) & 0.003 & 0.002 & 0.003 & 0.0067 & 0.0167 & 0.01 \\
Acc.\ truncation (m) & 0.017 & 0.01 & 0.017 & 0.033 & 0.083 & 0.1 \\
Comp.\ truncation (m) & 0.05 & 0.03 & 0.05 & 0.1 & 0.25 & 0.5 \\
Normal threshold ($^\circ$) & 30 & 30 & 30 & 30 & 30 & 30 \\
Sample density (pts/m$^2$) & $1\mathrm{e}{6}$ & $2.27\mathrm{e}{6}$ & $1\mathrm{e}{6}$ & -- & -- & -- \\
Crop volume & \checkmark & \checkmark & \checkmark & \checkmark & \checkmark & -- \\
\bottomrule
\end{tabular}
\caption{\textbf{Per-scene training and evaluation parameters.} Read from the configuration each run wrote into its own output directory. Shared by all ten methods on that scene; parameters that differ between methods are in \cref{tab:tandt_method_params}. Conventions: \cref{sec:appendix_full_params}.}
\label{tab:tandt_scene_params}
\end{table*}

\begin{table*}[t]
\centering
\scriptsize
\setlength{\tabcolsep}{4pt}
\begin{tabular}{llrrrrrr}
\toprule
Method & Quantity & Truck & Ignatius & Caterpillar & Barn & Courthouse & Redwood \\
\midrule
\multirow{1}{*}{Base} & Learning rate & 0.01 & 0.01 & 0.01 & 0.01 & 0.01 & 0.01 \\
\midrule
\multirow{2}{*}{RegSDF} & Learning rate & 0.01 & 0.01 & 0.01 & 0.01 & 0.01 & 0.01 \\
 & Weight & 0.003 & 0.003 & 0.003 & 0.003 & 0.003 & 0.003 \\
\midrule
\multirow{2}{*}{NeuralSSD} & Learning rate & 0.01 & 0.01 & 0.01 & 0.01 & 0.01 & 0.01 \\
 & Weight & $3\mathrm{e}{-5}$ & $3\mathrm{e}{-5}$ & $3\mathrm{e}{-5}$ & $3\mathrm{e}{-5}$ & $3\mathrm{e}{-5}$ & $3\mathrm{e}{-5}$ \\
\midrule
\multirow{2}{*}{GradHess} & Learning rate & 0.01 & 0.01 & 0.01 & 0.01 & 0.01 & 0.01 \\
 & Weight & $3\mathrm{e}{-7}$ & $3\mathrm{e}{-7}$ & $3\mathrm{e}{-7}$ & $3\mathrm{e}{-7}$ & $3\mathrm{e}{-7}$ & $3\mathrm{e}{-7}$ \\
\midrule
\multirow{2}{*}{Singular Hessian} & Learning rate & 0.01 & 0.01 & 0.01 & 0.01 & 0.01 & 0.01 \\
 & Weight & $1\mathrm{e}{-4}$ & $1\mathrm{e}{-4}$ & $1\mathrm{e}{-4}$ & $1\mathrm{e}{-4}$ & $1\mathrm{e}{-4}$ & $1\mathrm{e}{-4}$ \\
\midrule
\multirow{2}{*}{DiGS} & Learning rate & 0.01 & 0.01 & 0.01 & 0.01 & 0.01 & 0.01 \\
 & Weight & $3\mathrm{e}{-4}$ & $3\mathrm{e}{-4}$ & $3\mathrm{e}{-4}$ & $3\mathrm{e}{-4}$ & $3\mathrm{e}{-4}$ & $3\mathrm{e}{-4}$ \\
\midrule
\multirow{3}{*}{SIREN} & Learning rate & 0.01 & 0.01 & 0.01 & 0.01 & 0.01 & 0.01 \\
 & Weight & $3\mathrm{e}{-4}$ & $3\mathrm{e}{-4}$ & $3\mathrm{e}{-4}$ & $3\mathrm{e}{-4}$ & $3\mathrm{e}{-4}$ & $3\mathrm{e}{-4}$ \\
 & Activation & sine & sine & sine & sine & sine & sine \\
\midrule
\multirow{2}{*}{DiffCD} & Learning rate & 0.01 & 0.01 & 0.01 & 0.01 & 0.01 & 0.01 \\
 & Weight & 0.1 & 0.1 & 0.1 & 0.1 & 0.1 & 0.1 \\
\midrule
\multirow{2}{*}{Stoch. Precond.} & Learning rate & 0.01 & 0.01 & 0.01 & 0.01 & 0.01 & 0.01 \\
 & $\sigma_{\mathrm{sp}}$ (m) & 0.08 & 0.05 & 0.1 & 0.1 & 0.1 & 0.05 \\
\midrule
\multirow{3}{*}{\textbf{Ours}} & Learning rate & 0.001 & 0.01 & 0.01 & 0.01 & 0.01 & 0.01 \\
 & Weight & $3.12\mathrm{e}{-6}$ & $1.279\mathrm{e}{-6}$ & $5.546\mathrm{e}{-6}$ & $3.972\mathrm{e}{-5}$ & $1.693\mathrm{e}{-4}$ & $1.748\mathrm{e}{-5}$ \\
\bottomrule
\end{tabular}
\caption{\textbf{Per-method parameters.} The only quantities that differ between methods on a scene; the rest are in \cref{tab:tandt_scene_params}. Each method enables one regularizer on the common backbone. Each weight is the setting that maximized F-score in the sweep of \cref{fig:ablation_wsweep}, applied unchanged to every scene.}
\label{tab:tandt_method_params}
\end{table*}

\end{document}